\pdfoutput=1
\documentclass[11pt]{article}

\usepackage[preprint]{acl}

\usepackage{times}
\usepackage{latexsym}
\usepackage{colortbl}
\usepackage[T1]{fontenc}
\usepackage[utf8]{inputenc}

\usepackage{microtype}

\usepackage{inconsolata}

\usepackage{graphicx}

\usepackage{latexsym}
\usepackage{multirow}
\usepackage{booktabs}
\usepackage{subfigure}
\usepackage{amsfonts}
\usepackage{amsthm}

\usepackage{tablefootnote}
\usepackage{float}
\usepackage{enumitem}
\usepackage{algpseudocode}
\usepackage{amsmath}
\usepackage{tabularx} 
\usepackage{graphicx}
\usepackage{lipsum}
\usepackage{cuted}
\usepackage[linesnumbered, ruled, vlined]{algorithm2e}
\usepackage{amssymb}
\title{Dynamic Task Vector Grouping for Efficient Multi-Task Prompt Tuning}

\author{
 \textbf{Peiyi Zhang\textsuperscript{1}},
 \textbf{Richong Zhang\textsuperscript{1,2}\thanks{Corresponding author}},
 \textbf{Zhijie Nie\textsuperscript{1,3}},
\\
 \textsuperscript{1}CCSE, School of Computer Science and Engineering, Beihang University, Beijing, China\\
 \textsuperscript{2}Zhongguancun Laboratory, Beijing, China\\
 \textsuperscript{3}Shen Yuan Honors College, Beihang University, Beijing, China
\\
 \tt \{zhangpy,zhangrc,niezj\}@act.buaa.edu.cn
}

\begin{document}
\maketitle
\begin{abstract}

Multi-task prompt tuning utilizes multiple high-resource source tasks to improve performance on low-source target tasks. Existing approaches transfer the soft prompt trained by combining all source tasks or a single ``high-similar'' source task one-time-only. However, we find that the optimal transfer performance often comes from a combination of source tasks, which is neither one nor all. Further, we find that the similarity between source and target tasks also changes dynamically during fine-tuning after transfering, making similarity calculation in the initiation stage inadequate. To address these issues, we propose a method called Dynamic Task Vector Grouping (DTVG), whose core ideas contain (1) measuring the task similarity with task vectors instead of soft prompt, (2) grouping the optimal source task combination based on two metrics: {\it target similarity} and {\it knowledge consistency}; (3) dynamically updating the combination in each iteration step. Extensive experiments on the 26 NLP datasets under different settings demonstrate that DTVG effectively groups similar source tasks while reducing negative transfer, achieving the start-of-art performance. 
% \footnote{Our code is available at \url{https://anonymous.4open.science/r/DTVG-CD4E}.}

\end{abstract}

\begin{figure}[t!]
    \centering
    \includegraphics[width=\linewidth]{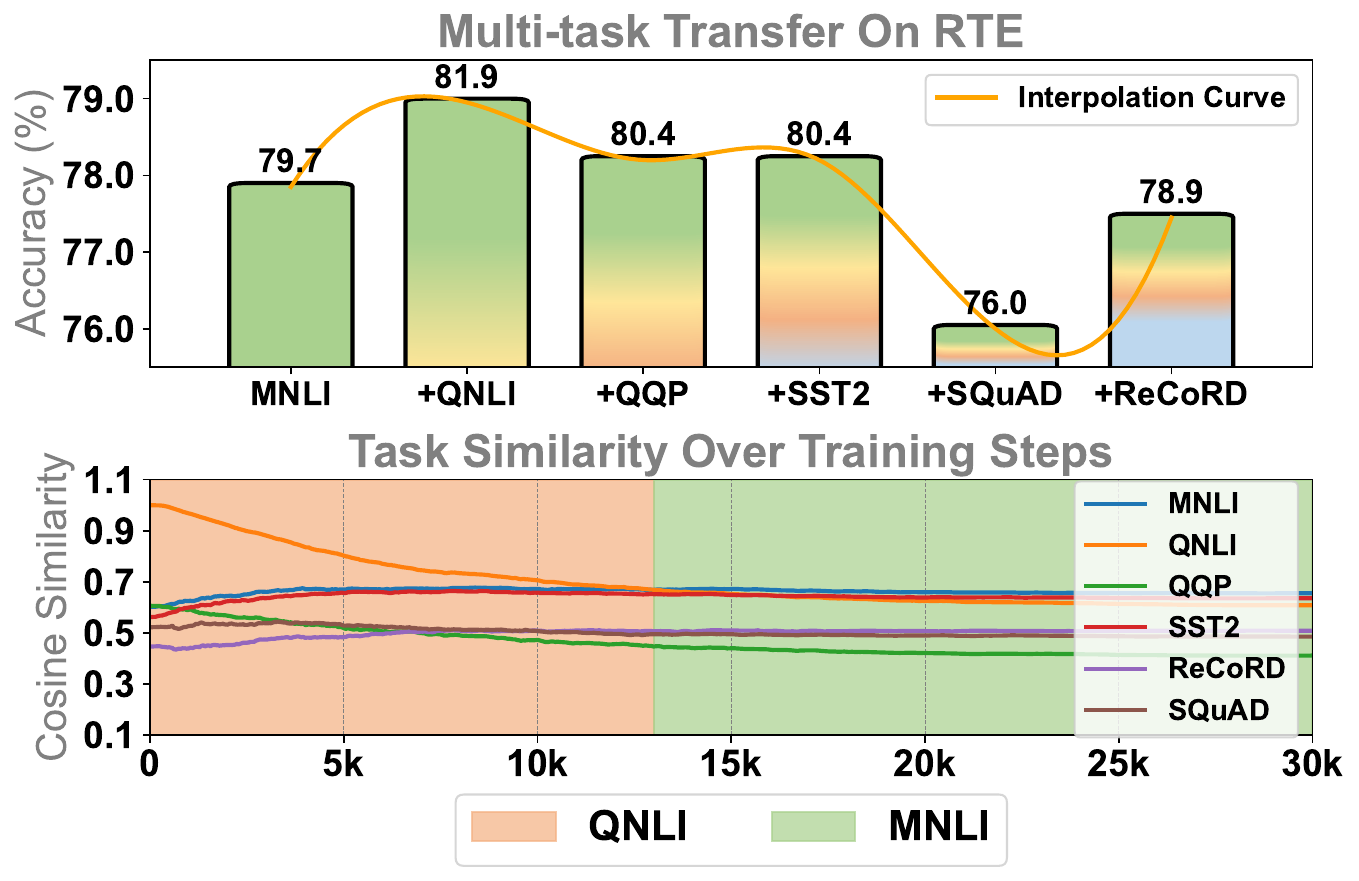}
    \caption{In the upper part, we use performance on the RTE validation set to study potential conflicts of source tasks. We incrementally add source tasks with a random order and train soft prompt by examples-proportional mixing~\cite{raffel2020exploring}. In the bottom part, we calculate the cosine similarity between the average pooled representations of the prompt tokens~\cite{vu2022spot}. We initialize the RTE soft prompt using the source task's soft prompt with the highest similarity. The legend marker denotes the source task with the highest similarity, which shifts from QNLI to MNLI during fine-tuning.}
    \label{fig:motivation}
\end{figure}

\section{Introduction}

Full parameter fine-tuning (FT) of large pre-trained language models (PLMs) has shown significant success in addressing various natural language processing (NLP) tasks. However, the conventional fine-tuning paradigm requires substantial memory and computational resources. Recently, parameter efficient fine-tuning (PEFT)~\citep{houlsby2019parameter,DBLP:conf/acl/LiL20,lester2021power,zaken2022bitfit,DBLP:conf/iclr/HuSWALWWC22} aims to achieve comparable results of FT by updating a significantly small set of the model parameters.

Soft prompt tuning (PT)~\cite{lester2021power}, as an effective PEFT method, achieves a trade-off between effectiveness and efficiency. During training, a series of learnable soft prompt vectors prepended to the input are updated while the original PLMs are frozen. Unlike methods such as LoRA~\cite{DBLP:conf/iclr/HuSWALWWC22} and Adapter~\cite{houlsby2019parameter}, PT is independent of the model architecture and can be applied to various models without modification. Although promising, existing study~\cite{asai2022attempt} demonstrates PT still underperforms compared to FT, particularly in the case of low-resource tasks. An additional issue with PT is sensitivity to the initialization and needs longer tuning for converge~\cite{lester2021power}.

Recent works~\citep{vu2022spot,asai2022attempt,feng2023learning,DBLP:conf/iclr/WangPKF0K23} address the above limitations by transferring soft prompt from high-resource source task to low-resource target task.

Specifically, they initialize the soft prompt for the target task by either (1) learning a common soft prompt across all source tasks or (2) learning a soft prompt for each source task and selecting one with the task similarity. Subsequently, the soft prompt is tuned exclusively using limited training samples from the target task. These transfer approaches effectively maintain the parameters efficiency of soft prompts and demonstrate superior performance compared to vanilla prompt tuning.

Despite substantial progress, we challenge the rationality of some straightforward ideas in existing approaches. We first check whether existing methods achieve optimal performance. In the upper part of Figure~\ref{fig:motivation}, we observe that  a subset of source tasks achieves the best transfer performance, neither all source tasks nor a source task. Additionally, \citet{vu2022spot} demonstrates MNLI, QNLI, and QQP positively transfer to the RTE dataset, while we find that their gradual addition does not yield a consistent monotonic improvement due to the potential conflicts among source tasks. These observations revealed that we should find a group of source tasks for each target task and consider potential conflicts between source tasks besides the similarity to target tasks.

Further, we check whether {\it ``the most similar source task''} will change in the tuning stage of the target task. We study a single-task version of SPoT~\cite{vu2022spot}, which transfers the soft prompt from a source task to initiate the target task via similarity measure between their learned soft prompt. In the bottom part of Figure~\ref{fig:motivation}, we find that {\it ``the most similar source task''} of RTE shifts from QNLI to MNLI over time. Recall that the low-resource characteristics of the target task hinder sufficient convergence of soft prompt; therefore, it is unsurprising that we cannot select the truly most similar task with an unconverted soft prompt of the target task. This observation suggests that dynamically updating the selected source task during the target task's fine-tuning may enhance the sustainable acquisition of knowledge.

Motivated by these valuable empirical observations, we propose a method called Dynamic Task Vector Grouping (DTVG). 
Specifically, We first introduce a novel task similarity metric, the dot product between task prompt vectors (TPV), which steadily achieves a better transfer performance than the current metric, the cosine similarity between soft prompts. Based on this metric, we introduce a source task grouping method to select the transfer source task group for each target task with two metrics, including {\it target similarity} and {\it knowledge consistency}. Then, a multi-task merging method is used to weighted sum the task vectors from the target task and the selected source tasks, synthesizing the initialization soft prompt for the target task. During the fine-tuning stage of the target task, we track the task similarity changes and dynamically update the source task group, which will effectively improve transfer performance.

In summary, our major contributions are to:
\begin{itemize}
    \item We present an effective task similarity metric, based on the task prompt vectors, to measure the transfer performance between tasks.
    \item We propose DTVG, a dynamic task vector grouping method that assembles and updates a source task group for each target task throughout the iterative training process to ensure sustainable acquisition of knowledge.
    \item We confirm the effectiveness of DTVG on the 26 datasets based on T5 and Llama3 under different settings, surpassing the advanced models and achieving SOTA performance.
\end{itemize} 
\section{Background}

\paragraph{Soft Prompt Tuning}

Soft Prompt Tuning (PT)~\citep{lester2021power} proposes strategically inserting the learned soft prompt into the input. Formally, for a task $t$ with the dataset $\mathcal{D}=\{(x_i, y_i)\}^{|D|}_{i=1}$, we fine-tuning a pre-trained model $F_{\Theta}$ to perform better in the task $t$ with its parameter $\Theta$ frozen. Instead, the learnable soft prompt $P \in \mathbb{R}^{d\times r}$ is introduced, where $d$ is the hidden state dimension of $F_{\Theta}$ and $r$ is the soft prompt length. The soft prompt $P$ and the token embedding matrix $E(x_i)$ are spliced as the input of $F_{\Theta}$. Then, the soft prompt $P^*$ is learned to boost the posterior probability of correct output $y_i$:
\begin{equation}
    P^* = \underset{P}{\arg\max}\  \mathbb{E}_{(x_i, y_i) \in \mathcal{D}}[\text{P}(y_i|P;E(x_i)])]
\end{equation}

Although the PT method has shown great success in various NLP tasks, it still faces the low-source challenge: Too few training samples prevent the soft prompts from converging, which can result in huge performance differences under different soft prompt initializations.

\begin{figure*}[tp]
    \centering
    \includegraphics[width=\textwidth]{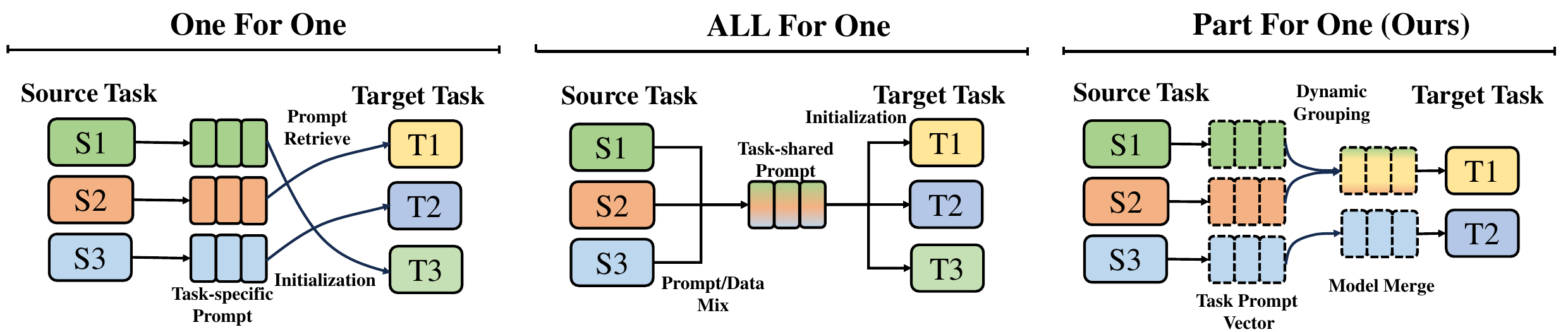}
    \caption{An overview of methods for comparison. One For One, initialize a target task by retrieving the task-specific prompt from one of the most similar source tasks based on task similarity. ALL For One, initialize a target task by learning appropriately across all source tasks based on prompt or data mix. Our Method: Part For One, dynamic group a subset of source tasks and merge their task prompt vectors.}
    \label{fig:comparison}
\end{figure*}

\paragraph{Multi-Task Prompt Tuning} Multi-Task Prompt Tuning~\cite{mahabadi2021parameter,vu2022spot,asai2022attempt,DBLP:conf/iclr/WangPKF0K23} is proposed to address the low-source challenge of PT. Formally, given a high-resource source task set $\mathcal{S}  = \{s^{1},s^{2},\dots,s^{n} \}$, where $n$ is the number of source tasks, Multi-Task Prompt Tuning improve the performance of a low-source target task $t$ by transfer learning from $\mathcal{S}$. Current methods usually contain two stages: (1) Learning the transferable soft prompts $P_{\rm mix}$ from $\mathcal{S}$, defined as $P_{\rm mix}=G(\mathcal{S},t)$ where $G$ is the learning method; (2) Adopting $P_{\rm mix}$ to $t$ and re-tuning $P_{\rm mix}$ with maximum training steps $N_{\rm max}$ on the training set of task $t$.

Multi-task Prompt Tuning does not impose restrictions on $G$ to get ${P_{\rm mix}}$  and how to adopt $P_{\rm mix}$ on $t$, excepting that the transfer ones must be soft prompts. Therefore, there are two representative lines of work to be highlighted. \underline{One For One}: $G$ serves as a retriever and selects the learned soft prompt of the most similar $s$ to initialize for $t$. SPoT~\cite{vu2022spot} regards the soft prompts as the task embeddings and measures task similarity via cosine similarity between soft prompts. \citet{feng2023learning} learns $G$ to predict transfer gain by randomly sampling soft prompt pairs. \underline{All For One}: $G$ serves as a blender and learns the task-shared prompt from source task set $\mathcal{S}$ via different mix strategies. SPoT~\cite{vu2022spot} also learns a single soft prompt through multi-task learning by mixing data. ATTEMPT~\cite{asai2022attempt} trains an attention module and mixes instance-wise prompts from all source tasks $\mathcal{S}$. MPT~\cite{DBLP:conf/iclr/WangPKF0K23} extends the multi-task training method of SPoT by learning task-shared and task-specific modules. TPT~\cite{DBLP:conf/emnlp/WuLXLLLHZH23} propose to retrieve token-wise soft prompt from the prompt bank.

\paragraph{Task Arithmetic} Task Arithmetic~\cite{DBLP:conf/iclr/IlharcoRWSHF23,zhang2024knowledge,ortiz2024task} as a newly emerged cost-effective approach demonstrates the effectiveness of multi-task training by operating task vectors derived from different tasks, where task vectors are given as the relative difference between the initialized parameters and those obtained after fine-tuning, capturing the changes induced by the adaptation process in weight space. Our proposed approach is inspired by Task Arithmetic. Similar to task vectors, the task prompt vectors (TPV) $T=[v_1,\dots,v_r] \in \mathbb{R}^{d\times r} $ are defined as the difference between $P_{\rm init}$ and $P^{*}$, i.e. $T = P^{*}-P_{\rm init}$, where $v_{i} \in \mathbb{R}^{d \times 1}$ represent the $i$-th vector in $T$. Concurrent work~\cite{belanec2024task} uses TPV to enable generalization to new target tasks without training. In contrast, we introduce TPV to address the issue of potential negative transfer in multi-task prompt tuning.

\begin{figure*}[tp]
    \centering
    \includegraphics[width=\textwidth]{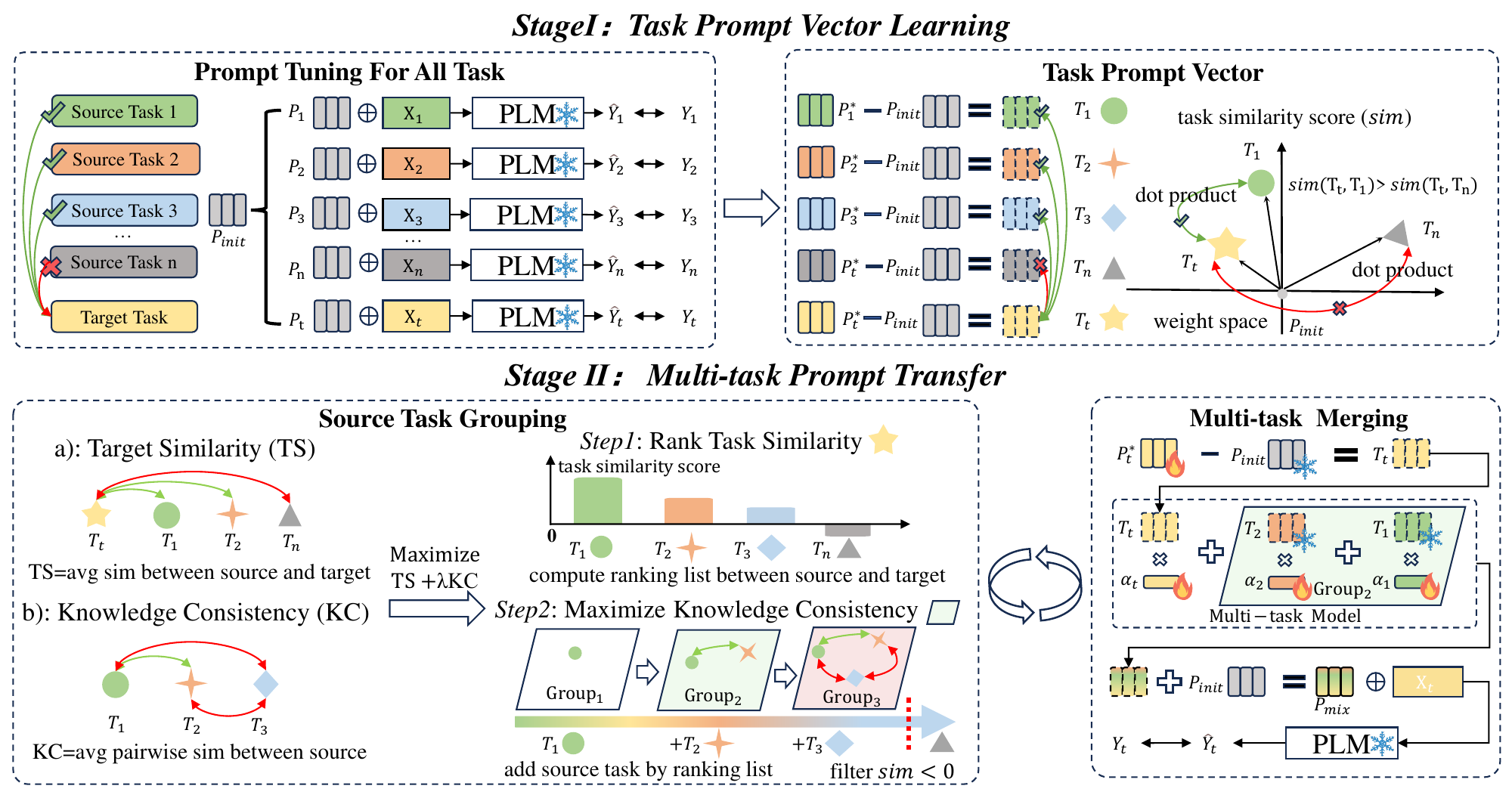}
    \caption{DTVG is to learn dynamic grouping partially related source tasks, including two stages: I) Task prompt vector Learning; II) Multi-task Prompt Transfer. In the first stage, we obtain task prompt vectors via vanilla prompt tuning. In the second stage, Source Task Grouping and Multi-task Merging are executed at each iteration step.}
    \label{fig:model}
\end{figure*}

\section{Method}

\subsection{Overview}
We propose a novel multi-task prompt tuning approach, Dynamic Task Vector Grouping (DTVG), which dynamically groups a subset from the source task set to transfer to the target task. Therefore, $G$ in our method serves as a grouper, allowing a specific target task to selectively leverage partially related source tasks, mitigating the risk of negative transfer. As shown in Figure~\ref{fig:comparison}, DTVG actually follows the idea of Part For One and distinguishes itself from existing methods.

DTVG consists of two stages: (\uppercase\expandafter{\romannumeral1}) Task Prompt Vector Learning to obtain a tuned TPV for each source and target task and (\uppercase\expandafter{\romannumeral2}) Multi-Task Prompt Transfer to group source tasks' TPV and merge it with the target vector's TPV. Note that the first stage only needs to be performed once, while the second stage is iterative, and the source task group will be dynamically updated during the fine-tuning process of the target task. An algorithm-style process of DTVG is provided in Appendix~\ref{sec:DTVG}.

\subsection{Task Prompt Vector Learning}
In the first stage, we obtain the soft prompt by individually tuning both the source and target task via the same initialization $P_{\rm init}$ and calculate their task prompt vectors $T$. Therefore, we have $n+1$ task prompt vectors from $\mathcal{S} \cup \{t\}$.

We propose using the average token-wise task prompt vectors to compute their dot product, allowing us to predict task similarity. This method enables a quantitative assessment of task relationships, as illustrated at the top of Figure~\ref{fig:model}. Specifically, given two task prompt vectors $T_1$ and $T_2$ from $s^1$ and $s^2$, we can calculate the similarity between tasks $s^1$ and $s^2$. The task similarity scores $sim$ between tasks is defined  as follows:
\begin{equation}\label{eqn:task_sim}
sim(T_1, T_2) = \frac{1}{r^2}\left(\sum_{i=1}^{r}v^{1}_{i}\right)^\top \left(\sum_{j=1}^{r}v^{2}_{j}\right)
\end{equation}
where $r$ denotes the length of soft prompt tokens.

To evaluate the effectiveness of this metric, we conduct transfer experiments on the SuperGLUE benchmark. As shown in Table \ref{tab:TPV}, TPV demonstrates consistent positive transfer, whereas SPoT exhibits negative transfer on WSC and CB, showing the superiority of our metric. Please refer to Appendix \ref{sec:TPV_SPoT} and \ref{sec:task_sim} for the experiment details and visual analysis, respectively.
\begin{table}[htb]
    \centering
    \resizebox{\linewidth}{!}{%
        \begin{tabular}{c|cccccc}
        \toprule
            & \multicolumn{6}{c}{\bf{SuperGLUE}}  \\
            \hline   
            \bf{Method} & Multi & Bool & WiC & WSC & CB & \bf{Avg.} \\
            \hline 
            PT & 72.7 & 76.0 & 62.6 & 67.3 & 82.1 & 72.1 \\
            SPoT & 74.9 \textcolor{green}{$\uparrow$} & 80.6 \textcolor{green}{$\uparrow$} &  65.2 \textcolor{green}{$\uparrow$} & 63.5 \textcolor{red}{$\downarrow$} & 78.6 \textcolor{red}{$\downarrow$} & 72.6 \\
            TPV & 74.2 \textcolor{green}{$\uparrow$} & 81.3 \textcolor{green}{$\uparrow$} & 66.1 \textcolor{green}{$\uparrow$} & 67.3 - & 92.9 \textcolor{green}{$\uparrow$} & 76.4  \\
        \hline
        \end{tabular}
    }
    \caption{Performance on SuperGLUE benchmark.}
    \label{tab:TPV}
\end{table}

\subsection{Multi-task Prompt Transfer}
In the second stage, we introduce an iterative process for multi-task prompt transfer. As shown in the bottom of Figure \ref{fig:model}, for each iteration, Source Task Grouping and Multi-Task Merging are executed sequentially to obtain $P_{\rm mix}$. 

\paragraph{Source Task Grouping}

Source task grouping aims to group a subset of source tasks $\mathcal{S}' \subseteq \mathcal{S}$. Source tasks in $\mathcal{S}'$ should not only be similar to the target task but also possess consistency of knowledge. We propose two metrics to characterize the source task group quantitatively, including {\it Target Similarity} and {\it Knowledge Consistency}.

    \underline{\bf Target Similarity}: To measure the transferability of multiple source tasks to the target task, we define a target similarity score $\rm TS$ as the average of the similarity between each source and target task prompt vector pair $(T_i, T_t)$, which is formulated as
    
    \begin{equation}
    {\rm TS}(\mathcal{S},t) =\frac{1}{|{\mathcal S}|} \sum_{s^i \in \mathcal{S}} sim(T_i, T_t)
    \end{equation}
    
    \underline{\bf Knowledge Consistency}: In multi-task transfer learning scenarios, conflicts among source tasks are prevalent. For example, in NLP, words crucial for sentiment (e.g., ``good'') may have varying significance in topic classification, leading to ambiguity and reduced performance in the target task. We propose to quantify the conflicts within a task group by calculating the average pairwise $sim$ between tasks. More formally, we defined the Knowledge Consistency Score ($\rm KC$): 
\begin{equation}
          {\rm KC} = 
        \begin{cases}
            \frac{2}{n (n - 1)} \sum\limits_{i < j} sim(T_i, T_j) & \text{if } |\mathcal{S}| \geq 2, \\
            0 & \text{otherwise}
        \end{cases}  
\end{equation}

    Therefore, the objective for selecting a source task group can be defined using $TS$ and $KC$:
    \begin{equation}
    \max_{\mathcal{S}' \subseteq \mathcal{S}} \left({\rm TS}(\mathcal{S}', t) + \lambda {\rm KC}(\mathcal{S}') \right) \label{obj:NP}
    \end{equation}
    \[
    \text{subject to}, \quad \forall s^i, s^j \in \mathcal{S}', \; s^i \neq s^j
    \]
    where $\lambda$ is the hyperparameter to achieve the trade-off between $\rm TS$ and $\rm KC$.
    
    However, the process to find the optimal $S'$ is equivalent to the Set Cover problem, which is the NP-Hard. As the number of tasks increases, the selection from $2^{|\mathcal{S}|}$ subsets becomes infeasible. Therefore, we use a heuristic algorithm to find a suboptimal subset, achieving a balance between efficiency and effectiveness. 
    % We propose the selection of a suboptimal subset to serve as a substitute for the exact solution, thereby achieving a balance between efficiency and effectiveness. 
    As shown in Figure \ref{fig:model}, the algorithm consists of two steps: (1)  $sim(T_i, T_t)$ between each source task $s^i$ and target task $t$ is computed. Then $\{sim(T_i, T_t)\}_{i=1}^n$ is ranked in the descending order, obtaining a rank list $\Pi = \{\pi^1,\pi^2,\dots,\pi^n\}$, where $\pi^j$ is the source task with $j$-th highest $sim(T_i, T_t)$; (2) The source tasks with $sim(T_i, T_t)$ are added to the set $S'$ one by one in the order in T until the ${\rm KC}(\mathcal{S}')$ is no longer increasing. The implementation details are provided in the Appendix~\ref{sec:STG}.

\paragraph{Multi-task  Merging}
Multi-task merging aims to merge the task prompt vectors from the source task group and the target task to get a final soft prompt $P_{\rm mix}$. Specifically, $P_{\rm mix}$ is obtained by the sum of (1) the rescaled soft prompt task vectors of $\mathcal{S}' \cup \{t\}$ and (2) a common initialization prompt, which can be denoted as
\begin{equation}
    P_{\rm mix} = \underbrace{P_{init} \vphantom{\alpha_t T_{t}  + \sum_{s \in \mathcal{S}'}\tau_sT_{s}}}_{\text{Initialization}} 
    \mathrel{+} \underbrace{\alpha_t T_{t}  + \sum_{s \in \mathcal{S}'}\alpha_sT_{s}}_{\text{Merged Task Prompt Vector}}
\end{equation}
where $\alpha  \in \mathbb{R}^{l}$ is token level scaling term initialized to all-ones vector. In practice, we employ rescaled task prompt vectors to compute the task similarity score (Equation~\ref{eqn:task_sim}).

\paragraph{Iteration Update}
In each training step, we sequentially execute the above two steps to ensure the correct source task group $\mathcal{S}'$ selection to compute $P_{\rm mix}$ with in-batch. In practice, we observe that in the early stages of training, the grouping of source tasks exhibits significant fluctuations due to the insufficient convergence of the target task prompt vectors. As the iterations progress, the dynamic grouping gradually stabilizes and ultimately maintains consistency (see Section \ref{sec:additional_analysis} for details).

\section{Experiments}

\subsection{Experiment Setup}
\paragraph{Datasets}
We evaluate the model's natural language understanding capabilities using the GLUE and SuperGLUE benchmarks. In addition, we also use four question-answering datasets from the MRQA 2019 benchmark and four datasets from the ``other'' benchmark. In the following, we introduce the source tasks and target tasks separately. Further details can be found in Appendix~\ref{sec:Dataset}.

    \underline{\bf{Source Tasks}}: Following \citet{DBLP:conf/iclr/WangPKF0K23}, we set $n$ to 6, and use the same large-scale datasets as source tasks, including MNLI, QNLI, QQP, SST2 from GLUE, ReCoRD from SuperGLUE, and SQuAD from MRQA 2019. 
    
     \underline{\bf{Target Tasks}}: we use all 8 datasets from GLUE, 5 datasets (excluding 
 ReCoRD) from SuperGLUE, 4 datasets (excluding SQuAD) from MRQA 2019, and 4 datasets from the ``other'' benchmark.

\paragraph{Models}
We adopt the model setup from \cite{lester2021power} for prompt tuning. Our experiments mainly utilize T5-base with the soft prompt of length 100, while in ablation studies, we also explore other scales of T5 in Section \ref{sec:additional_analysis}.

\paragraph{Baselines} We compare our method with several baseline methods. (1) no transfer learning, which updates model parameters for the each target task without source task, including Finetuning (FT), Prompt Tuning (PT)~\cite{lester2021power}, BitFit~\cite{zaken2022bitfit}, Adapter~\cite{houlsby2019parameter}, LoRA~\cite{DBLP:conf/iclr/HuSWALWWC22}, DePT~\cite{DBLP:conf/iclr/ShiL24}, as well as multi-task versions of FT, Adapter, HyperFomer~\cite{mahabadi2021parameter}, and HyperDecoder~\cite{ivison2022hyperdecoders}. Note that we exclude ACCEPT~\cite{lin2024accept} due to the lack of accessible open-source code, which prevents an evaluation of its ability to address sensitivity to prompt initialization.
 (2) transfer learning + one for one, where transfer soft prompt from one source task to each target task, such as SPoT \cite{vu2022spot} (3) transfer learning + all for one, where transfer soft prompt from all source tasks to each target task, including ATTEMPT~\cite{asai2022attempt}, MPT~\cite{DBLP:conf/iclr/WangPKF0K23}, TPT~\cite{DBLP:conf/emnlp/WuLXLLLHZH23} as well as multi-task versions of ATTEMPT, and MPT. For a fair comparison, we directly quote the results of the baselines reported in previous works~\citep{asai2022attempt,DBLP:conf/emnlp/WuLXLLLHZH23,DBLP:conf/iclr/WangPKF0K23,DBLP:conf/iclr/ShiL24} whenever possible, and utilize publicly available source code to ensure consistent experimental settings.

\paragraph{Implementation Details} For both the Task Prompt Vector Learning and Multi-task Prompt Transfer stage, we train on high-resource source tasks for 300K steps, following \citet{vu2022spot}. For the target tasks, we set $N_{\rm max}$ to 30K. Aligning with standard prompt tuning methods \cite{lester2021power}, we use a default learning rate of 0.3 and select checkpoints with the highest validation set scores to extract task prompt vectors. In the Multi-task Prompt Transfer stage, we apply two-speed learning rates for different modules. We conduct transfer experiments four times and report the average results. Please see Appendix~\ref{sec:Implementation} for details.

\paragraph{Parameter Efficiency}
For both source and target tasks, we compute the task prompt vector $T \in \mathbb{R}^{r*d}$, where $r$ is the length of the soft prompt and $d$ is the model dimension. For each source task, we introduce a learned scaling term $\alpha \in \mathbb{R}^{r}$. Our framework enables knowledge transfer from partial source tasks to the target task, therefore, the total number of learned parameters ranges from $r+r*d = r*(d+1)$ to $(n+1)*r+r*d = r*(d+n+1)$, where $n$ is the number of source tasks. We compare different methods' trainable parameters under the least favorable conditions of DTVG in Table \ref{tab:GLUE_SuperGLUE}. 

\subsection{Main Results}
\paragraph{Full-dataset Transfer}

Table~\ref{tab:GLUE_SuperGLUE} provides the performance and parameter comparison for each dataset on the GLUE and SuperGLUE benchmarks across different baselines. Additionally, we visualize the result on GLUE (see Appendix~\ref{sec:vis_glue_superglue}).
Notably, our proposed method, DTVG, outperforms others by achieving the {\it highest average performance} on GLUE and SuperGLUE with a {\it minimal parameter tuning fraction} of $0.035\%$, in contrast to the fine-tuning. When compared to prompt tuning in terms of low-resource datasets, DTVG significantly improves the performance of the target task, such as CoLA (10.6\% vs. 69.1\%) and CB (67.9\% vs. 97.6\%). Simultaneously, our multiple experiments demonstrate that DTVG is robust for addressing inappropriate soft prompt initialization leading to performance degradation. Please see Appendix~\ref{sec:MRQA_other_Benchmark} for details on MRQA and ``Other'' benchmarks.

% gary
\definecolor{lightblue}{rgb}{0.85, 0.85, 0.85}  % RGB: (188, 211, 232)
% blue
\definecolor{lightgray}{rgb}{0.8515625, 0.90625, 0.95703125}  % RGB: (188, 211, 232)

\begin{table*}[ht]
  \centering
  \tabcolsep=0.014cm
  \small
  \begin{tabular}{llccccccccc|cccccc}
    \hline
    & &  \multicolumn{9}{|c|}{\bf{GLUE}} & \multicolumn{6}{c}{\bf{SuperGLUE}} \\
    \hline

    \multirow{2}{*}{\textbf{Method}} & \multirow{2}{*}{ \shortstack{\bf{param} \\ $\backslash$ \bf{task}} } & 
    \cellcolor{lightblue} MNLI & \cellcolor{lightblue}QQP & \cellcolor{lightblue} QNLI & \cellcolor{lightblue} SST2 & 
    \cellcolor{lightgray}STS-B & \cellcolor{lightgray}MRPC & \cellcolor{lightgray}RTE & \cellcolor{lightgray} CoLA & \multirow{2}{*}{\textbf{Avg.}} & \cellcolor{lightgray}Multi & \cellcolor{lightgray} Bool & \cellcolor{lightgray}WiC & \cellcolor{lightgray}WSC & \cellcolor{lightgray} CB   & \multirow{2}{*}{\textbf{Avg.}} \\
    & & \cellcolor{lightblue}(393K) & \cellcolor{lightblue}(364K) & \cellcolor{lightblue}(105K) & \cellcolor{lightblue}(67K) & \cellcolor{lightgray}(7K) & \cellcolor{lightgray}(3.7K) & \cellcolor{lightgray}(2.5K) & \cellcolor{lightgray}(8.5K) &  & \cellcolor{lightgray}(5.1K) & \cellcolor{lightgray}(9.4K) & \cellcolor{lightgray}(6K) & \cellcolor{lightgray}(554) & \cellcolor{lightgray}(250) \\
    
        \hline
        \multicolumn{17}{c}{\textit{\textbf{no transfer learning}}} \\
        \hline
        Finetuning$_{1}$ & \multicolumn{1}{c|}{220M} & 86.8 & 91.6 & 93.0 & 94.6 & 89.7 & 90.2 & 71.9 & 61.8 & 84.9 & 72.8 & 81.1 & 70.2 & 59.6 & 85.7 & 73.9 \\
        PT$_{1}$ & \multicolumn{1}{c|}{76.8K} & 81.3 & 89.7 & 92.8 & 90.9 & 89.5 & 68.1 & 54.7 & 10.6 & 72.2 & 58.7 & 61.7 & 48.9 & 51.9 & 67.9 & 57.8 \\
        BitFit$_{1}$ & \multicolumn{1}{c|}{280K} & 85.3 & 90.1 & 93.0 & 94.2 & 90.9 & 86.8 & 67.6 & 58.2 & 83.3 & 74.5 & 79.6 & 70.0 & 59.6 & 78.6 & 72.5 \\
        Adapter$_{1}$ & \multicolumn{1}{c|}{1.9M} & \bf{86.5} & 90.2 & 93.2 & 93.8 & 90.7 & 85.3 & 71.9 & 64.0 & 84.5 & \textbf{75.9} & \textbf{82.5} & 67.1 & 67.3 & 85.7 & 75.7 \\
        LoRA$_{4}$ & \multicolumn{1}{c|}{3.8M} & 86.3 & 89.0 & 93.2 & 94.3 & 90.9 & 90.1 & 75.5 & 63.3 & 85.3 & 72.6 & 81.3 & 68.3 & 67.3 & 92.9 &76.5 \\
        DePT$_{4}$ & \multicolumn{1}{c|}{76.8k} & 85.0 & 90.4 & 93.2 & 94.2 & 90.8 & \textbf{90.7} & 79.1 & 63.8 & 85.9 & 74.3 & 79.3 & 68.7 & 67.3 & 92.9 & 76.5 \\
        Finetuning$^{\ast}_{1}$ & \multicolumn{1}{c|}{28M} & 85.7 & 91.1 & 92.0 & 92.5 & 88.8 & 90.2 & 75.4 & 54.9 & 83.8 & 74.4 & 81.1 & 70.0 & 71.2 & 85.7 & 76.1\\
        Adapters$^{\ast}_{1}$ & \multicolumn{1}{c|}{1.8M} & 86.3 & \textbf{90.5} & 93.2 & 93.0 & 89.9 & 90.2 & 70.3 & 61.5 & 84.4 & 72.6 & 82.3 & 66.5 & 67.3 & 89.3 & 75.6 \\
        HyperFomer$^{\ast}_{1}$ & \multicolumn{1}{c|}{638K} & 85.7 & 90.0 & 93.0 & 94.0 & 89.7 & 87.2 & 75.4 & 63.7 & 84.8 & 72.9 & 82.5 & 69.0 & 67.3 & 85.7 & 75.4 \\
        HyperDecoder$^{\ast}_{1}$ & \multicolumn{1}{c|}{1.8M} & 86.0 & \textbf{90.5} & \textbf{93.4} & 94.0 & 90.5 & 87.7 & 71.7 & 55.9 & 83.7 & 70.4 & 78.8 & 67.1 & 61.5 & 82.1 & 72.0\\
        \hline
        \multicolumn{17}{c}{\textit{\textbf{transfer learning + one for one}}} \\
        \hline
        SPoT$_{1}$ & \multicolumn{1}{c|}{76.8K} & 85.4 & 90.1 & 93.0 & 93.4 & 90.0 & 79.7 & 69.8 & 57.1 & 82.3 & 74.0 & 77.2 & 67.0 & 50.0 & 46.4 & 62.9 \\
        \hline
        \multicolumn{17}{c}{\textit{\textbf{transfer learning + all for one}}} \\
        \hline
        ATTEMPT$_{1}$ & \multicolumn{1}{c|}{232K} & 84.3 & 90.3 & 93.0 & 93.2 & 89.7 & 85.7 & 73.4 & 57.4 & 83.4 & 74.4 & 78.8 & 66.8 & 53.8 & 78.6 & 70.5 \\
        MPT$_{3}$ & \multicolumn{1}{c|}{77.6K} & 85.9 & 90.3 & 93.1 & 93.8 & 90.4 & 89.1 & 79.4 & 62.4 & 85.6 & 74.8 & 79.6 & 69.0 & 67.3 & 79.8 & 74.1 \\
        TPT$_{2}$ & \multicolumn{1}{c|}{539K} & 85.5 & 90.1 & 93.2 & \textbf{94.7} & 89.8 & 89.7 & 82.3 & 59.8 & 85.6 & 74.4 & 80.1 & 69.8 & 67.3 & 94.6 & 77.2 \\
        ATTEMPT$^{\ast}_{1}$ & \multicolumn{1}{c|}{96K} & 83.8 & 90.0 & 93.1 & 93.7 & 90.8 & 86.1 & 79.9 & 64.3 & 85.2 & 74.4 & 78.3 & 66.5 & 69.2 & 82.1 & 74.1 \\
        MPT$^{\ast}_{3}$ & \multicolumn{1}{c|}{10.5K} & 84.3  & 90.0 & 93.0 & 93.3 & 90.4 & 89.2 & 82.7 & 63.5 & 85.8 & 74.8 & 79.2 & 70.2 & 67.3 & 89.3 & 76.1 \\
        \hline
        \multicolumn{17}{c}{\textit{\textbf{transfer learning + part for one}}} \\
        \hline
        DTVG (ours) & \multicolumn{1}{c|}{77.5K} & 86.0\textsubscript{0.2} & 90.3\textsubscript{0.1} & 93.1\textsubscript{0.0} & 93.2\textsubscript{0.0}
        & \textbf{91.0}\textsubscript{0.2} & 90.4\textsubscript{0.2} & \textbf{86.3}\textsubscript{0.6} & \textbf{69.1}\textsubscript{1.0} & \textbf{87.4} 
        & 74.5\textsubscript{0.7} & 81.4\textsubscript{0.1} & \textbf{71.1}\textsubscript{0.5} & \textbf{69.9}\textsubscript{3.6} & \textbf{97.6}\textsubscript{3.4} & \textbf{78.9} \\
        \hline
    \end{tabular}  
    
    \caption{Results on GLUE and SuperGLUE benchmark.
    ``param$\backslash$task`` denotes the number of learnable parameters for each task on the GLUE. * denotes multi-task learning on target tasks. $_{1}$ sourced from \citet{asai2022attempt}, $_{2}$ sourced from \citet{DBLP:conf/emnlp/WuLXLLLHZH23}, $_{3}$ sourced from \citet{DBLP:conf/iclr/WangPKF0K23} and $_{4}$ sourced from \citet{DBLP:conf/iclr/ShiL24}. We differentiate high-resource and low-resource tasks using gray and blue, respectively, to highlight our contribution.
    }
    \label{tab:GLUE_SuperGLUE}    
\end{table*}

\paragraph{Few-shot Adaptation}
We compare our method with other baselines on BoolQ, CB, and SciTail in Table \ref{tab:fs_BoolQ_CB_SciTail}. On average, our method outperforms the baselines in low-resource settings with only ($k$= 4,16,32) shots, indicating that our DTVG is adept at harnessing knowledge from multiple source tasks for effective transfer in scenarios with limited training samples. More details about GLUE and SuperGLUE are given in Appendix~\ref{sec:Few-shot GLUE_SuperGLUE}.

\begin{table}[t]
    \centering
    \resizebox{\linewidth}{!}{%
    \begin{tabular}{c|c|cccccc|c}
    \toprule
        \multirow{2}{*}{\textbf{Task}} & \multirow{2}{*}{$\bf{k}$} & \multicolumn{7}{c}{\textbf{Method}} \\ 
        &  & FT & PT & HF & ATP & MPT & DePT & Our \\
    \midrule
        ~ & \multicolumn{1}{|c|}{4} & 50.5 & 61.6 & 48.0 & 61.8 & 62.2 & \textbf{62.7}\textsubscript{5.4} & 60.6\textsubscript{1.5} \\ 
        BoolQ & \multicolumn{1}{|c|}{16} & 56.5 & 61.9 & 50.2 & 60.0 & 63.3 & 66.9\textsubscript{4.4} & \textbf{72.3}\textsubscript{1.4} \\ 
        ~ & \multicolumn{1}{|c|}{32} & 58.4 & 61.7 & 58.3 & 65.3 & 68.9 & 67.2\textsubscript{3.4} & \textbf{73.5}\textsubscript{1.1} \\
        \midrule
        ~ & \multicolumn{1}{|c|}{4} & 57.7 & 53.5 & 60.7 & \textbf{82.1} & 73.6 & 75.0\textsubscript{5.1} & \textbf{86.9}\textsubscript{1.7} \\
        CB & \multicolumn{1}{|c|}{16} & 77.0 & 63.5 & 76.3 & 78.5 & 78.6 & 78.6\textsubscript{4.3} & \textbf{82.1}\textsubscript{2.9} \\ 
        ~ & \multicolumn{1}{|c|}{32} & 80.0 & 67.8 & 81.4 & \textbf{85.7} & 82.1 & 82.1\textsubscript{2.3} & 84.5\textsubscript{1.7} \\ 
        \midrule
        ~ & \multicolumn{1}{|c|}{4} & 79.6 & 57.7 & \textbf{82.0} & 80.2 & 80.2 & 78.1\textsubscript{2.5} & 78.3\textsubscript{1.1} \\ 
        SciTail & \multicolumn{1}{|c|}{16} & 80.0 & 60.8 & 86.5 & 79.5 & \textbf{87.3} & 78.5\textsubscript{1.4} & 82.1\textsubscript{2.9} \\ 
        ~ & \multicolumn{1}{|c|}{32} & 81.9 & 60.2 & 85.8 & 80.2 & \textbf{86.3} & 85.4\textsubscript{3.1} & 85.3\textsubscript{2.5} \\ 
    \bottomrule
    \end{tabular}}
    \caption{Few-shot adaptation on BoolQ, CB, and SciTail datasets, where FT, HF, ATP denote Finetuning, HyperFomer, and ATTEMPT, respectively.}
    \label{tab:fs_BoolQ_CB_SciTail}
\end{table}

\subsection{Ablation Study}

\begin{figure}[t]
    \centering
    \begin{subfigure}
        \centering
        \captionsetup{skip=2pt}
        \includegraphics[width=\linewidth]{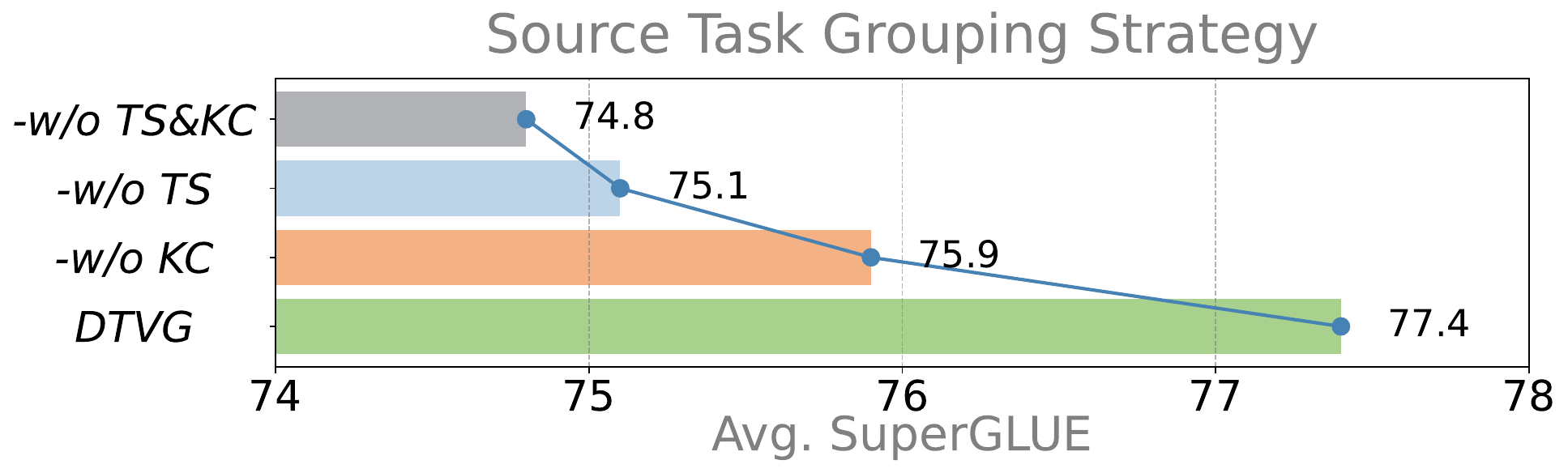} 
        \caption{Ablation study for the source task grouping.} 
        \label{fig:ablation_two_perspectives}
    \end{subfigure}
 
    \hspace{-8pt} 
    \begin{subfigure}
        \centering
        \captionsetup{skip=2pt} 
        \includegraphics[width=\linewidth]{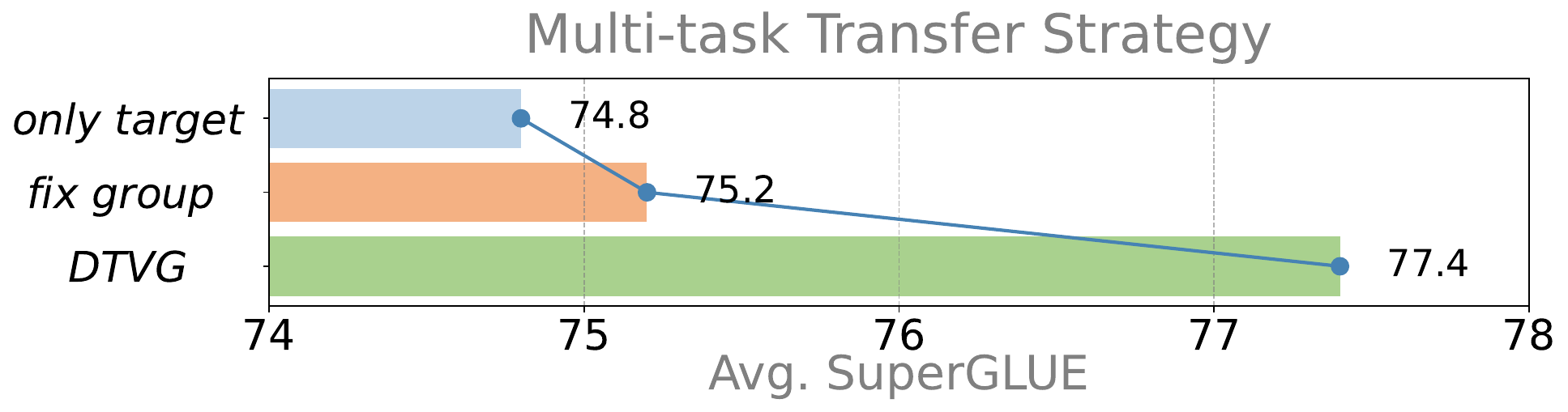} 
        \caption{Ablation study for multi-task prompt transfer.} 
        \label{fig:ablation_transfer_strategy} 
    \end{subfigure}
    \label{fig:combined_figures}
\end{figure}

\paragraph{Source Task Grouping Strategy}
We conduct ablation experiments on the SuperGLUE benchmark to study the impact of two different perspectives for source task grouping. For a) Target Similarity (TS), we only merge TPV with $sim\geq0$. For b) Knowledge Consistency (KC), we select the source task group with the highest $KC$ among all source task combinations. As shown in Figure~\ref{fig:ablation_two_perspectives}, these strategies can improve performance consistently. KC improves the average performance on SuperGLUE from 74.8 to 75.1, suggesting that mitigating the conflict among multiple source tasks is critical for effective multi-task prompt tuning, even when the task combinations may not be directly related to the target tasks. ST improves the average performance on SuperGLUE from 74.8 to 75.9, indicating that $sim$ can effectively evaluate and leverage similar source tasks for transferring.

\paragraph{Multi-task Prompt Transfer Strategy} We conduct a study to ablate different multi-task prompt transfer strategies, including 1) {\it only target}: This strategy focuses solely on learning the target task prompt and its associated scaling term for the task prompt vectors. 2) {\it fix group}: This strategy fixes the initial source task group, thus eliminating the effect of dynamic grouping, which relies on the specific grouping of source tasks.
Figure~\ref{fig:ablation_transfer_strategy} shows that using a fixed group of source tasks
results in a performance drop (77.4 vs. 75.2), suggesting that the choice of the source task group is important. This emphasizes the need for our approach, namely DTVG’s ability to efficiently group source tasks by dynamic iteration, thereby improving performance.

\subsection{Additional Analysis}\label{sec:additional_analysis}

We extend our experiments to comprehensively evaluate the performance of DTVG, including model scaling, natural language generation, generalization to other LLMs, and dynamic grouping during training. However, for some experiments without a standard evaluation protocol, we analyze DTVG only against some fundamental baselines.

\begin{figure}[tb]
    \centering
    \includegraphics[width=\linewidth]{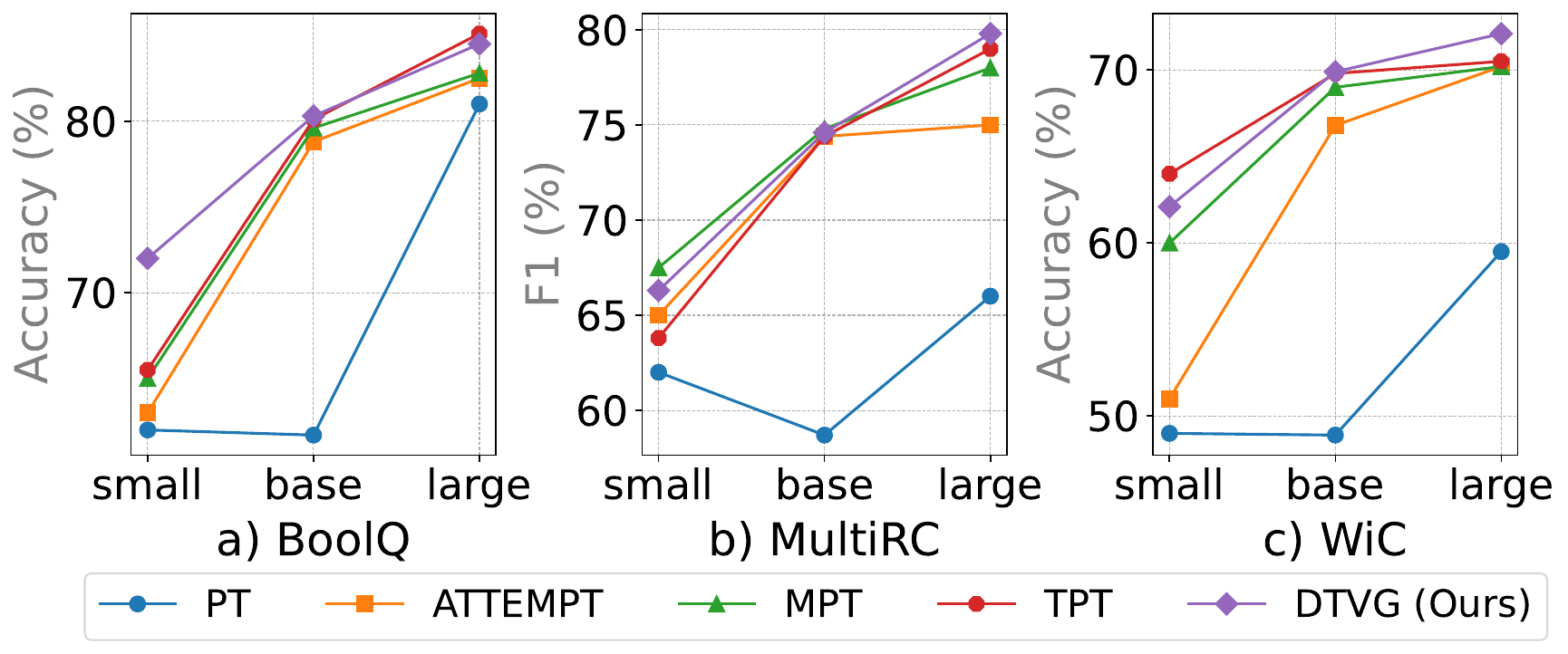}
    \caption{Model Scaling on BoolQ, MultiRC, and WiC.}
    \label{fig:ablation_model_scale}
\end{figure}

\paragraph{Model Scaling} Figure~\ref{fig:ablation_model_scale} illustrates the results on three SuperGLUE datasets with different scales of the T5 model. We observe that as the model size increases, performance across different tasks improves. This indicates that our method indeed benefits from a larger model capacity. Please refer to Appendix~\ref{sec:model_scale} for experiment details.

\paragraph{Natural Language Generation} As shown in Table~\ref{tab:NLG_tasks}, we observe that DTVG consistently outperforms PT on three natural language generation tasks (namely, E2E~\cite{dusek-etal-2019-semantic}, CommonGen~\cite{DBLP:conf/emnlp/LinZSZBCR20}, and WebNLG~\cite{DBLP:conf/inlg/GardentSNP17}), suggesting DTVG works not only for NLU but also for NLG. Interestingly, although we transfer TPV from NLU tasks to NLG tasks, DTVG's performance on NLG tasks does not degrade, which aligns with the same observation~\cite{DBLP:conf/iclr/WangPKF0K23}. We suspect that this phenomenon might be related to T5's text-to-text framework. Please see Appendix~\ref{sec:other_LLMs} for details.
\begin{table}[t]
    \centering
    \resizebox{\linewidth}{!}{%
    \begin{tabular}{ll|cccc}
    \hline
        \multirow{2}{*}{\textbf{Task}} &
        \multirow{2}{*}{\textbf{Method}} &
        \multicolumn{4}{c}{\textbf{Metris}} \\
        & & BLEU  & R-1 & R-2 & R-L\\
    \hline
    \multirow{2}{*}{E2E}       
    & PT  & 0.274 	& 62.1	& 36.3	& 47.0 \\
    & DTVG  & 0.331 & 63.6 & 37.5 & 47.9 \\
    \hline
    \multirow{2}{*}{CommonGen} 
    & PT  & 0.056  & 33.3 & 9.9 & 27.6 \\
    & DTVG  & 0.067 &  36.6 & 11.0 & 29.1 \\
    \hline
    \multirow{2}{*}{WebNLG}   
    & PT    & 0.293 & 64.4 & 39.6 & 52.3 \\
    & DTVG  & 0.363 & 66.3 & 41.4 & 53.4 \\
    \hline
    \end{tabular}}
    \caption{Performance on NLG tasks. R-1, R-2, and R-L denote Rouge-1, Rouge-2, and Rouge-L, respectively.}
    \label{tab:NLG_tasks}
\end{table}

\paragraph{Generalization to Other LLMs} We experimentally analyze the performance of DTVG on the latest decoder-based models using Llama-3.2-1B and Llama-3.2-3B~\cite{dubey2024llama}. As shown in Table~\ref{tab:Other_LLMs}, DTVG outperforms vanilla prompt tuning across various target tasks. When compared with SPoT, DTVG demonstrates consistent positive transfer across various LLMs, whereas SPoT exhibits negative transfer, such as on RTE with Llama-3.2-1B (74.8\% vs. 57.6\%). These results suggest that DTVG's generalizability to other types of LLMs. Moreover, we observe that DTVG performs better on Llama-3.2-3B than Llama-3.2-1B, indicating that it benefits from more powerful LLMs. Please see Appendix~\ref{sec:nlg_tasks} for experiment details.

\begin{table}[t]
    \centering
    \resizebox{\linewidth}{!}{
    \begin{tabular}{c|cccc|cccc}
    \hline
      \multirow{2}{*}{\textbf{Method}}  & \multicolumn{4}{c|}{\textbf{LLama-3.2-1B}} & \multicolumn{4}{c}{\textbf{LLama-3.2-3B}} \\
      ~ & RTE & CoLA & CB & WSC & RTE & CoLA & CB & WSC\\
    \hline
    PT   & 74.8 & 59.2 & 60.7 & 63.5 & 60.4 & 67.2 & 64.3 & 67.3\\
    SPoT & 57.6 & 67.5 & 64.3 & 67.3 & 63.3 & 71.7 & 60.7 & 67.3 \\
    DTVG & 84.1 & 63.4 & 82.1 & 67.3  & 89.2 & 73.1 & 89.3 & 69.2 \\
    \hline
    \end{tabular}}
    \caption{Results on Llama-3.2-1B and Llama-3.2-3B.}
    \label{tab:Other_LLMs}
\end{table}

\begin{figure}[t]  
    \centering    
    \includegraphics[width=\linewidth]{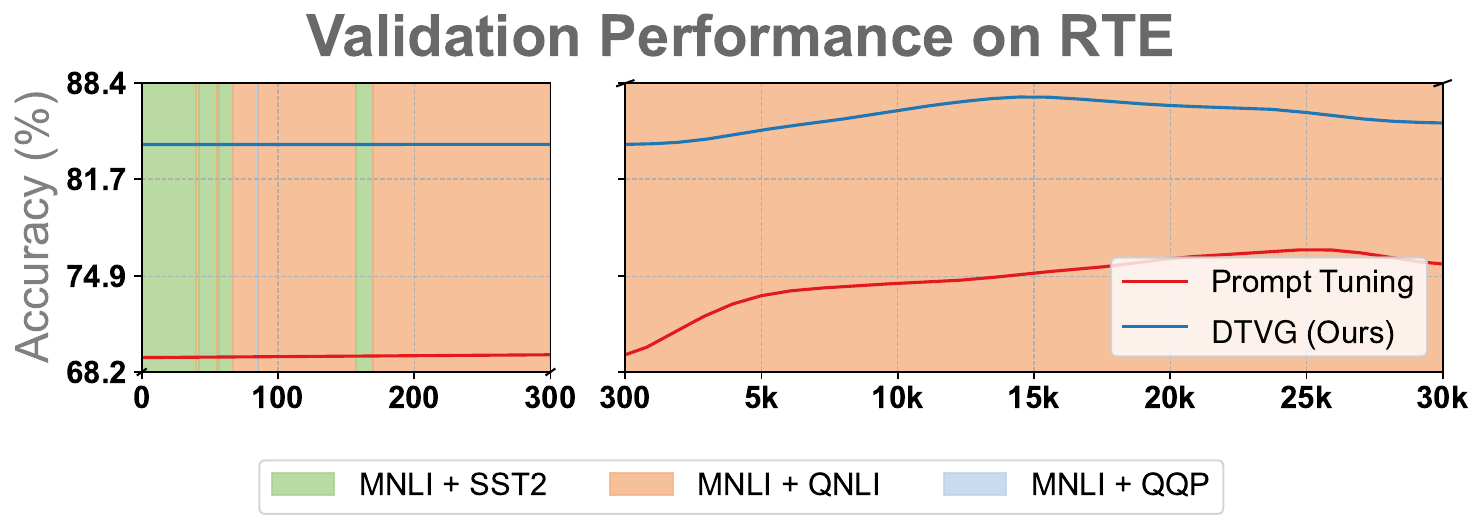} 
    \caption{Validation performance on RTE with source task grouping. The source tasks are arranged in each patch legend from left to right, ordered by their similarity to the target task, from highest to lowest. } 
    \label{fig:RTE}
\end{figure}

\paragraph{Dynamic Grouping} Figure~\ref{fig:RTE} illustrates the variations of dynamic grouping for RTE during the training process. Compared to prompt tuning, DTVG achieves better performance on RTE.

From the task grouping perspective, we observe the source task combination shifts from [top1: MNLI, top2: SST2] to [top1: MNLI, top2: QNLI] over time. This result suggests that a) {\textbf{Target Similarity}}: two NLI source tasks become more aligned to the target task RTE (NLI); and b) {\textbf{Knowledge Consistency}}: conflicts exist between MNLI and SST2 (replaced by QNLI) are reduced.

From an iterative training perspective, we observe that source task groups fluctuate frequently during the early stages of training. As training progresses, the task group converges, resulting in a stable selection of tasks in the final stage. This supports our hypothesis that insufficient convergence is attributed to the low-resource characteristics of the target tasks. Additionally, we report the grouping results of MRPC, NQ, and SciTail in Appendix~\ref{sec:TG}.

\section{Conclusion}
In this paper, we present DTVG, a novel approach for addressing potential negative transfer in multi-task prompt tuning based on task prompt vectors. Compared to vanilla transfer of the soft prompt from all source tasks, we dynamically group a subset of source tasks and merge their task prompt vectors to avoid an unrelated source task inducing performance degradation of the target task. Extensive experiments demonstrate that DTVG effectively groups related source tasks to further optimize the performance of the target task.

\section{Limitations}
DTVG has three limitations, 1) prompt length: although \citet{lester2021power} found that setting the soft prompt length to 100 achieved relatively strong results across various LLMs, the impact of varying prompt lengths on DTVG remains unknown. 2) computation and time costs: while our methods achieve superior performance on various NLU and NLG tasks, dynamically calculating the task combinations during each parameter update introduces additional computational and time burden. We investigate this issue in Appendix~\ref{sec:computation_time_cost}. A potential optimization is to perform source task grouping~\ref{sec:STG} every K step, and we will explore the feasible maximum value of K in the future. 3) source task number: we use 6 high-resource tasks to serve as source tasks, while the impact of the number of source tasks remains unknown and we will add more high-resource source tasks to evaluate the performance of DTVG.

\bibliography{custom}

\begin{thebibliography}{49}
\providecommand{\natexlab}[1]{#1}

\bibitem[{Asai et~al.(2022)Asai, Salehi, Peters, and Hajishirzi}]{asai2022attempt}
Akari Asai, Mohammadreza Salehi, Matthew~E Peters, and Hannaneh Hajishirzi. 2022.
\newblock Attempt: Parameter-efficient multi-task tuning via attentional mixtures of soft prompts.
\newblock In \emph{Proceedings of the 2022 Conference on Empirical Methods in Natural Language Processing}, pages 6655--6672.

\bibitem[{Belanec et~al.(2024)Belanec, Ostermann, Srba, and Bielikova}]{belanec2024task}
Robert Belanec, Simon Ostermann, Ivan Srba, and Maria Bielikova. 2024.
\newblock Task prompt vectors: Effective initialization through multi-task soft-prompt transfer.
\newblock \emph{arXiv preprint arXiv:2408.01119}.

\bibitem[{Cer et~al.(2017)Cer, Diab, Agirre, Lopez-Gazpio, and Specia}]{cer2017semeval}
Daniel Cer, Mona Diab, Eneko Agirre, I{\~n}igo Lopez-Gazpio, and Lucia Specia. 2017.
\newblock Semeval-2017 task 1: Semantic textual similarity multilingual and crosslingual focused evaluation.
\newblock In \emph{Proceedings of the 11th International Workshop on Semantic Evaluation (SemEval-2017)}, pages 1--14.

\bibitem[{Clark et~al.(2019)Clark, Lee, Chang, Kwiatkowski, Collins, and Toutanova}]{clark2019boolq}
Christopher Clark, Kenton Lee, Ming-Wei Chang, Tom Kwiatkowski, Michael Collins, and Kristina Toutanova. 2019.
\newblock Boolq: Exploring the surprising difficulty of natural yes/no questions.
\newblock \emph{arXiv preprint arXiv:1905.10044}.

\bibitem[{De~Marneffe et~al.(2019)De~Marneffe, Simons, and Tonhauser}]{de2019commitmentbank}
Marie-Catherine De~Marneffe, Mandy Simons, and Judith Tonhauser. 2019.
\newblock The commitmentbank: Investigating projection in naturally occurring discourse.
\newblock In \emph{proceedings of Sinn und Bedeutung}, pages 107--124.

\bibitem[{Demszky et~al.(2018)Demszky, Guu, and Liang}]{demszky2018transforming}
Dorottya Demszky, Kelvin Guu, and Percy Liang. 2018.
\newblock Transforming question answering datasets into natural language inference datasets.
\newblock \emph{arXiv preprint arXiv:1809.02922}.

\bibitem[{Dolan and Brockett(2005)}]{dolan2005automatically}
Bill Dolan and Chris Brockett. 2005.
\newblock Automatically constructing a corpus of sentential paraphrases.
\newblock In \emph{Third international workshop on paraphrasing (IWP2005)}.

\bibitem[{Dubey et~al.(2024)Dubey, Jauhri, Pandey, Kadian, Al-Dahle, Letman, Mathur, Schelten, Yang, Fan et~al.}]{dubey2024llama}
Abhimanyu Dubey, Abhinav Jauhri, Abhinav Pandey, Abhishek Kadian, Ahmad Al-Dahle, Aiesha Letman, Akhil Mathur, Alan Schelten, Amy Yang, Angela Fan, et~al. 2024.
\newblock The llama 3 herd of models.
\newblock \emph{arXiv preprint arXiv:2407.21783}.

\bibitem[{Dunn et~al.(2017)Dunn, Sagun, Higgins, Guney, Cirik, and Cho}]{dunn2017searchqa}
Matthew Dunn, Levent Sagun, Mike Higgins, V~Ugur Guney, Volkan Cirik, and Kyunghyun Cho. 2017.
\newblock Searchqa: A new q\&a dataset augmented with context from a search engine.
\newblock \emph{arXiv preprint arXiv:1704.05179}.

\bibitem[{Du{\v{s}}ek et~al.(2019)Du{\v{s}}ek, Howcroft, and Rieser}]{dusek-etal-2019-semantic}
Ond{\v{r}}ej Du{\v{s}}ek, David~M. Howcroft, and Verena Rieser. 2019.
\newblock \href {https://doi.org/10.18653/v1/W19-8652} {Semantic noise matters for neural natural language generation}.
\newblock In \emph{Proceedings of the 12th International Conference on Natural Language Generation}, pages 421--426, Tokyo, Japan. Association for Computational Linguistics.

\bibitem[{Feng(2023)}]{feng2023learning}
Lingyun Feng. 2023.
\newblock Learning to predict task transferability via soft prompt.
\newblock In \emph{Proceedings of the 2023 Conference on Empirical Methods in Natural Language Processing}, pages 8829--8844.

\bibitem[{Gardent et~al.(2017)Gardent, Shimorina, Narayan, and Perez{-}Beltrachini}]{DBLP:conf/inlg/GardentSNP17}
Claire Gardent, Anastasia Shimorina, Shashi Narayan, and Laura Perez{-}Beltrachini. 2017.
\newblock \href {https://doi.org/10.18653/V1/W17-3518} {The webnlg challenge: Generating text from {RDF} data}.
\newblock In \emph{Proceedings of the 10th International Conference on Natural Language Generation, {INLG} 2017, Santiago de Compostela, Spain, September 4-7, 2017}, pages 124--133. Association for Computational Linguistics.

\bibitem[{Gehrmann et~al.(2021)Gehrmann, Adewumi, Aggarwal, Ammanamanchi, Aremu, Bosselut, Chandu, Clinciu, Das, Dhole, Du, Durmus, Du{\v{s}}ek, Emezue, Gangal, Garbacea, Hashimoto, Hou, Jernite, Jhamtani, Ji, Jolly, Kale, Kumar, Ladhak, Madaan, Maddela, Mahajan, Mahamood, Majumder, Martins, McMillan-Major, Mille, van Miltenburg, Nadeem, Narayan, Nikolaev, Niyongabo~Rubungo, Osei, Parikh, Perez-Beltrachini, Rao, Raunak, Rodriguez, Santhanam, Sedoc, Sellam, Shaikh, Shimorina, Sobrevilla~Cabezudo, Strobelt, Subramani, Xu, Yang, Yerukola, and Zhou}]{gehrmann-etal-2021-gem}
Sebastian Gehrmann, Tosin Adewumi, Karmanya Aggarwal, Pawan~Sasanka Ammanamanchi, Anuoluwapo Aremu, Antoine Bosselut, Khyathi~Raghavi Chandu, Miruna-Adriana Clinciu, Dipanjan Das, Kaustubh Dhole, Wanyu Du, Esin Durmus, Ond{\v{r}}ej Du{\v{s}}ek, Chris~Chinenye Emezue, Varun Gangal, Cristina Garbacea, Tatsunori Hashimoto, Yufang Hou, Yacine Jernite, Harsh Jhamtani, Yangfeng Ji, Shailza Jolly, Mihir Kale, Dhruv Kumar, Faisal Ladhak, Aman Madaan, Mounica Maddela, Khyati Mahajan, Saad Mahamood, Bodhisattwa~Prasad Majumder, Pedro~Henrique Martins, Angelina McMillan-Major, Simon Mille, Emiel van Miltenburg, Moin Nadeem, Shashi Narayan, Vitaly Nikolaev, Andre Niyongabo~Rubungo, Salomey Osei, Ankur Parikh, Laura Perez-Beltrachini, Niranjan~Ramesh Rao, Vikas Raunak, Juan~Diego Rodriguez, Sashank Santhanam, Jo{\~a}o Sedoc, Thibault Sellam, Samira Shaikh, Anastasia Shimorina, Marco~Antonio Sobrevilla~Cabezudo, Hendrik Strobelt, Nishant Subramani, Wei Xu, Diyi Yang, Akhila Yerukola, and Jiawei Zhou. 2021.
\newblock \href {https://doi.org/10.18653/v1/2021.gem-1.10} {The {GEM} benchmark: Natural language generation, its evaluation and metrics}.
\newblock In \emph{Proceedings of the 1st Workshop on Natural Language Generation, Evaluation, and Metrics (GEM 2021)}, pages 96--120, Online. Association for Computational Linguistics.

\bibitem[{Giampiccolo et~al.(2007)Giampiccolo, Magnini, Dagan, and Dolan}]{giampiccolo2007third}
Danilo Giampiccolo, Bernardo Magnini, Ido Dagan, and William~B Dolan. 2007.
\newblock The third pascal recognizing textual entailment challenge.
\newblock In \emph{Proceedings of the ACL-PASCAL workshop on textual entailment and paraphrasing}, pages 1--9.

\bibitem[{Houlsby et~al.(2019)Houlsby, Giurgiu, Jastrzebski, Morrone, De~Laroussilhe, Gesmundo, Attariyan, and Gelly}]{houlsby2019parameter}
Neil Houlsby, Andrei Giurgiu, Stanislaw Jastrzebski, Bruna Morrone, Quentin De~Laroussilhe, Andrea Gesmundo, Mona Attariyan, and Sylvain Gelly. 2019.
\newblock Parameter-efficient transfer learning for nlp.
\newblock In \emph{International conference on machine learning}, pages 2790--2799. PMLR.

\bibitem[{Hu et~al.(2022)Hu, Shen, Wallis, Allen{-}Zhu, Li, Wang, Wang, and Chen}]{DBLP:conf/iclr/HuSWALWWC22}
Edward~J. Hu, Yelong Shen, Phillip Wallis, Zeyuan Allen{-}Zhu, Yuanzhi Li, Shean Wang, Lu~Wang, and Weizhu Chen. 2022.
\newblock \href {https://openreview.net/forum?id=nZeVKeeFYf9} {Lora: Low-rank adaptation of large language models}.
\newblock In \emph{The Tenth International Conference on Learning Representations, {ICLR} 2022, Virtual Event, April 25-29, 2022}. OpenReview.net.

\bibitem[{Ilharco et~al.(2023)Ilharco, Ribeiro, Wortsman, Schmidt, Hajishirzi, and Farhadi}]{DBLP:conf/iclr/IlharcoRWSHF23}
Gabriel Ilharco, Marco~T{\'{u}}lio Ribeiro, Mitchell Wortsman, Ludwig Schmidt, Hannaneh Hajishirzi, and Ali Farhadi. 2023.
\newblock \href {https://openreview.net/forum?id=6t0Kwf8-jrj} {Editing models with task arithmetic}.
\newblock In \emph{The Eleventh International Conference on Learning Representations, {ICLR} 2023, Kigali, Rwanda, May 1-5, 2023}. OpenReview.net.

\bibitem[{Ivison and Peters(2022)}]{ivison2022hyperdecoders}
Hamish Ivison and Matthew~E Peters. 2022.
\newblock Hyperdecoders: Instance-specific decoders for multi-task nlp.
\newblock In \emph{Findings of the Association for Computational Linguistics: EMNLP 2022}, pages 1715--1730.

\bibitem[{Khashabi et~al.(2018)Khashabi, Chaturvedi, Roth, Upadhyay, and Roth}]{khashabi2018looking}
Daniel Khashabi, Snigdha Chaturvedi, Michael Roth, Shyam Upadhyay, and Dan Roth. 2018.
\newblock Looking beyond the surface: A challenge set for reading comprehension over multiple sentences.
\newblock In \emph{Proceedings of the 2018 Conference of the North American Chapter of the Association for Computational Linguistics: Human Language Technologies, Volume 1 (Long Papers)}, pages 252--262.

\bibitem[{Khot et~al.(2018)Khot, Sabharwal, and Clark}]{khot2018scitail}
Tushar Khot, Ashish Sabharwal, and Peter Clark. 2018.
\newblock Scitail: A textual entailment dataset from science question answering.
\newblock In \emph{Proceedings of the AAAI conference on artificial intelligence}.

\bibitem[{Kingma and Ba(2015)}]{DBLP:journals/corr/KingmaB14}
Diederik~P. Kingma and Jimmy Ba. 2015.
\newblock \href {http://arxiv.org/abs/1412.6980} {Adam: {A} method for stochastic optimization}.
\newblock In \emph{3rd International Conference on Learning Representations, {ICLR} 2015, San Diego, CA, USA, May 7-9, 2015, Conference Track Proceedings}.

\bibitem[{Kwiatkowski et~al.(2019)Kwiatkowski, Palomaki, Redfield, Collins, Parikh, Alberti, Epstein, Polosukhin, Devlin, Lee et~al.}]{kwiatkowski2019natural}
Tom Kwiatkowski, Jennimaria Palomaki, Olivia Redfield, Michael Collins, Ankur Parikh, Chris Alberti, Danielle Epstein, Illia Polosukhin, Jacob Devlin, Kenton Lee, et~al. 2019.
\newblock Natural questions: A benchmark for question answering research.
\newblock \emph{Transactions of the Association for Computational Linguistics}, 7:452--466.

\bibitem[{Lester et~al.(2021)Lester, Al-Rfou, and Constant}]{lester2021power}
Brian Lester, Rami Al-Rfou, and Noah Constant. 2021.
\newblock The power of scale for parameter-efficient prompt tuning.
\newblock In \emph{Proceedings of the 2021 Conference on Empirical Methods in Natural Language Processing}, pages 3045--3059.

\bibitem[{Levesque et~al.(2012)Levesque, Davis, and Morgenstern}]{levesque2012winograd}
Hector~J Levesque, Ernest Davis, and Leora Morgenstern. 2012.
\newblock The winograd schema challenge.
\newblock In \emph{Proceedings of the Thirteenth International Conference on Principles of Knowledge Representation and Reasoning}, pages 552--561.

\bibitem[{Li and Liang(2021)}]{DBLP:conf/acl/LiL20}
Xiang~Lisa Li and Percy Liang. 2021.
\newblock \href {https://doi.org/10.18653/V1/2021.ACL-LONG.353} {Prefix-tuning: Optimizing continuous prompts for generation}.
\newblock In \emph{Proceedings of the 59th Annual Meeting of the Association for Computational Linguistics and the 11th International Joint Conference on Natural Language Processing, {ACL/IJCNLP} 2021, (Volume 1: Long Papers), Virtual Event, August 1-6, 2021}, pages 4582--4597. Association for Computational Linguistics.

\bibitem[{Lin et~al.(2020)Lin, Zhou, Shen, Zhou, Bhagavatula, Choi, and Ren}]{DBLP:conf/emnlp/LinZSZBCR20}
Bill~Yuchen Lin, Wangchunshu Zhou, Ming Shen, Pei Zhou, Chandra Bhagavatula, Yejin Choi, and Xiang Ren. 2020.
\newblock \href {https://doi.org/10.18653/V1/2020.FINDINGS-EMNLP.165} {Commongen: {A} constrained text generation challenge for generative commonsense reasoning}.
\newblock In \emph{Findings of the Association for Computational Linguistics: {EMNLP} 2020, Online Event, 16-20 November 2020}, volume {EMNLP} 2020 of \emph{Findings of {ACL}}, pages 1823--1840. Association for Computational Linguistics.

\bibitem[{Lin et~al.(2024)Lin, Li, Chen, and Chen}]{lin2024accept}
Yu-Chen Lin, Wei-Hua Li, Jun-cheng Chen, and Chu-Song Chen. 2024.
\newblock Accept: Adaptive codebook for composite and efficient prompt tuning.
\newblock In \emph{Findings of the Association for Computational Linguistics: EMNLP 2024}, pages 15345--15358.

\bibitem[{Mahabadi et~al.(2021)Mahabadi, Ruder, Dehghani, and Henderson}]{mahabadi2021parameter}
Rabeeh~Karimi Mahabadi, Sebastian Ruder, Mostafa Dehghani, and James Henderson. 2021.
\newblock Parameter-efficient multi-task fine-tuning for transformers via shared hypernetworks.
\newblock In \emph{Proceedings of the 59th Annual Meeting of the Association for Computational Linguistics and the 11th International Joint Conference on Natural Language Processing (Volume 1: Long Papers)}, pages 565--576.

\bibitem[{Ortiz-Jimenez et~al.(2024)Ortiz-Jimenez, Favero, and Frossard}]{ortiz2024task}
Guillermo Ortiz-Jimenez, Alessandro Favero, and Pascal Frossard. 2024.
\newblock Task arithmetic in the tangent space: Improved editing of pre-trained models.
\newblock \emph{Advances in Neural Information Processing Systems}, 36.

\bibitem[{Phang et~al.(2018)Phang, F{\'e}vry, and Bowman}]{phang2018sentence}
Jason Phang, Thibault F{\'e}vry, and Samuel~R Bowman. 2018.
\newblock Sentence encoders on stilts: Supplementary training on intermediate labeled-data tasks.
\newblock \emph{arXiv preprint arXiv:1811.01088}.

\bibitem[{Pilehvar and Camacho-Collados(2019)}]{pilehvar2019wic}
Mohammad~Taher Pilehvar and Jose Camacho-Collados. 2019.
\newblock Wic: the word-in-context dataset for evaluating context-sensitive meaning representations.
\newblock In \emph{Proceedings of the 2019 Conference of the North American Chapter of the Association for Computational Linguistics: Human Language Technologies, Volume 1 (Long and Short Papers)}, pages 1267--1273.

\bibitem[{Raffel et~al.(2020)Raffel, Shazeer, Roberts, Lee, Narang, Matena, Zhou, Li, and Liu}]{raffel2020exploring}
Colin Raffel, Noam Shazeer, Adam Roberts, Katherine Lee, Sharan Narang, Michael Matena, Yanqi Zhou, Wei Li, and Peter~J Liu. 2020.
\newblock Exploring the limits of transfer learning with a unified text-to-text transformer.
\newblock \emph{Journal of machine learning research}, 21(140):1--67.

\bibitem[{Rajpurkar(2016)}]{rajpurkar2016squad}
P~Rajpurkar. 2016.
\newblock Squad: 100,000+ questions for machine comprehension of text.
\newblock \emph{arXiv preprint arXiv:1606.05250}.

\bibitem[{Sakaguchi et~al.(2021)Sakaguchi, Bras, Bhagavatula, and Choi}]{sakaguchi2021winogrande}
Keisuke Sakaguchi, Ronan~Le Bras, Chandra Bhagavatula, and Yejin Choi. 2021.
\newblock Winogrande: An adversarial winograd schema challenge at scale.
\newblock \emph{Communications of the ACM}, 64(9):99--106.

\bibitem[{Shi and Lipani(2024)}]{DBLP:conf/iclr/ShiL24}
Zhengxiang Shi and Aldo Lipani. 2024.
\newblock \href {https://openreview.net/forum?id=KjegfPGRde} {Dept: Decomposed prompt tuning for parameter-efficient fine-tuning}.
\newblock In \emph{The Twelfth International Conference on Learning Representations, {ICLR} 2024, Vienna, Austria, May 7-11, 2024}. OpenReview.net.

\bibitem[{Socher et~al.(2013)Socher, Perelygin, Wu, Chuang, Manning, Ng, and Potts}]{socher2013recursive}
Richard Socher, Alex Perelygin, Jean Wu, Jason Chuang, Christopher~D Manning, Andrew~Y Ng, and Christopher Potts. 2013.
\newblock Recursive deep models for semantic compositionality over a sentiment treebank.
\newblock In \emph{Proceedings of the 2013 conference on empirical methods in natural language processing}, pages 1631--1642.

\bibitem[{Trischler et~al.(2017)Trischler, Wang, Yuan, Harris, Sordoni, Bachman, and Suleman}]{trischler2017newsqa}
Adam Trischler, Tong Wang, Xingdi Yuan, Justin Harris, Alessandro Sordoni, Philip Bachman, and Kaheer Suleman. 2017.
\newblock Newsqa: A machine comprehension dataset.
\newblock In \emph{Proceedings of the 2nd Workshop on Representation Learning for NLP}, pages 191--200.

\bibitem[{Vu et~al.(2022)Vu, Lester, Constant, Al-Rfou, and Cer}]{vu2022spot}
Tu~Vu, Brian Lester, Noah Constant, Rami Al-Rfou, and Daniel Cer. 2022.
\newblock Spot: Better frozen model adaptation through soft prompt transfer.
\newblock In \emph{Proceedings of the 60th Annual Meeting of the Association for Computational Linguistics (Volume 1: Long Papers)}, pages 5039--5059.

\bibitem[{Wang(2018)}]{wang2018glue}
Alex Wang. 2018.
\newblock Glue: A multi-task benchmark and analysis platform for natural language understanding.
\newblock \emph{arXiv preprint arXiv:1804.07461}.

\bibitem[{Wang et~al.(2023)Wang, Panda, Karlinsky, Feris, Sun, and Kim}]{DBLP:conf/iclr/WangPKF0K23}
Zhen Wang, Rameswar Panda, Leonid Karlinsky, Rog{\'{e}}rio Feris, Huan Sun, and Yoon Kim. 2023.
\newblock \href {https://openreview.net/forum?id=Nk2pDtuhTq} {Multitask prompt tuning enables parameter-efficient transfer learning}.
\newblock In \emph{The Eleventh International Conference on Learning Representations, {ICLR} 2023, Kigali, Rwanda, May 1-5, 2023}. OpenReview.net.

\bibitem[{Warstadt(2019)}]{warstadt2019neural}
A~Warstadt. 2019.
\newblock Neural network acceptability judgments.
\newblock \emph{arXiv preprint arXiv:1805.12471}.

\bibitem[{Williams et~al.(2018)Williams, Nangia, and Bowman}]{williams2018broad}
Adina Williams, Nikita Nangia, and Samuel Bowman. 2018.
\newblock A broad-coverage challenge corpus for sentence understanding through inference.
\newblock In \emph{Proceedings of the 2018 Conference of the North American Chapter of the Association for Computational Linguistics: Human Language Technologies, Volume 1 (Long Papers)}, pages 1112--1122.

\bibitem[{Wu et~al.(2023)Wu, Liu, Xu, Lv, Ling, Li, Huang, Zheng, and Huang}]{DBLP:conf/emnlp/WuLXLLLHZH23}
Muling Wu, Wenhao Liu, Jianhan Xu, Changze Lv, Zixuan Ling, Tianlong Li, Longtao Huang, Xiaoqing Zheng, and Xuanjing Huang. 2023.
\newblock \href {https://doi.org/10.18653/V1/2023.FINDINGS-EMNLP.584} {Parameter efficient multi-task fine-tuning by learning to transfer token-wise prompts}.
\newblock In \emph{Findings of the Association for Computational Linguistics: {EMNLP} 2023, Singapore, December 6-10, 2023}, pages 8734--8746. Association for Computational Linguistics.

\bibitem[{Yang et~al.(2018)Yang, Qi, Zhang, Bengio, Cohen, Salakhutdinov, and Manning}]{yang2018hotpotqa}
Zhilin Yang, Peng Qi, Saizheng Zhang, Yoshua Bengio, William Cohen, Ruslan Salakhutdinov, and Christopher~D Manning. 2018.
\newblock Hotpotqa: A dataset for diverse, explainable multi-hop question answering.
\newblock In \emph{Proceedings of the 2018 Conference on Empirical Methods in Natural Language Processing}, pages 2369--2380.

\bibitem[{Zaken et~al.(2022)Zaken, Goldberg, and Ravfogel}]{zaken2022bitfit}
Elad~Ben Zaken, Yoav Goldberg, and Shauli Ravfogel. 2022.
\newblock Bitfit: Simple parameter-efficient fine-tuning for transformer-based masked language-models.
\newblock In \emph{Proceedings of the 60th Annual Meeting of the Association for Computational Linguistics (Volume 2: Short Papers)}, pages 1--9.

\bibitem[{Zhang et~al.(2024)Zhang, Albert, Rodriguez-Opazo, Hengel, and Abbasnejad}]{zhang2024knowledge}
Frederic~Z Zhang, Paul Albert, Cristian Rodriguez-Opazo, Anton van~den Hengel, and Ehsan Abbasnejad. 2024.
\newblock Knowledge composition using task vectors with learned anisotropic scaling.
\newblock \emph{arXiv preprint arXiv:2407.02880}.

\bibitem[{Zhang et~al.(2018)Zhang, Liu, Liu, Gao, Duh, and Van~Durme}]{zhang2018record}
Sheng Zhang, Xiaodong Liu, Jingjing Liu, Jianfeng Gao, Kevin Duh, and Benjamin Van~Durme. 2018.
\newblock Record: Bridging the gap between human and machine commonsense reading comprehension.
\newblock \emph{arXiv preprint arXiv:1810.12885}.

\bibitem[{Zhang et~al.(2015)Zhang, Zhao, and LeCun}]{zhang2015character}
Xiang Zhang, Junbo Zhao, and Yann LeCun. 2015.
\newblock Character-level convolutional networks for text classification.
\newblock \emph{Advances in neural information processing systems}, 28.

\bibitem[{Zhang et~al.(2019)Zhang, Baldridge, and He}]{zhang2019paws}
Yuan Zhang, Jason Baldridge, and Luheng He. 2019.
\newblock Paws: Paraphrase adversaries from word scrambling.
\newblock In \emph{Proceedings of the 2019 Conference of the North American Chapter of the Association for Computational Linguistics: Human Language Technologies, Volume 1 (Long and Short Papers)}, pages 1298--1308.

\end{thebibliography}

\clearpage
\appendix
{\noindent\large\textbf{Appendix}}\label{sec:appendix}

\section{Performance and Parameter Comparison}

\begin{figure}[htb]
    \centering
    \includegraphics[width=\linewidth]{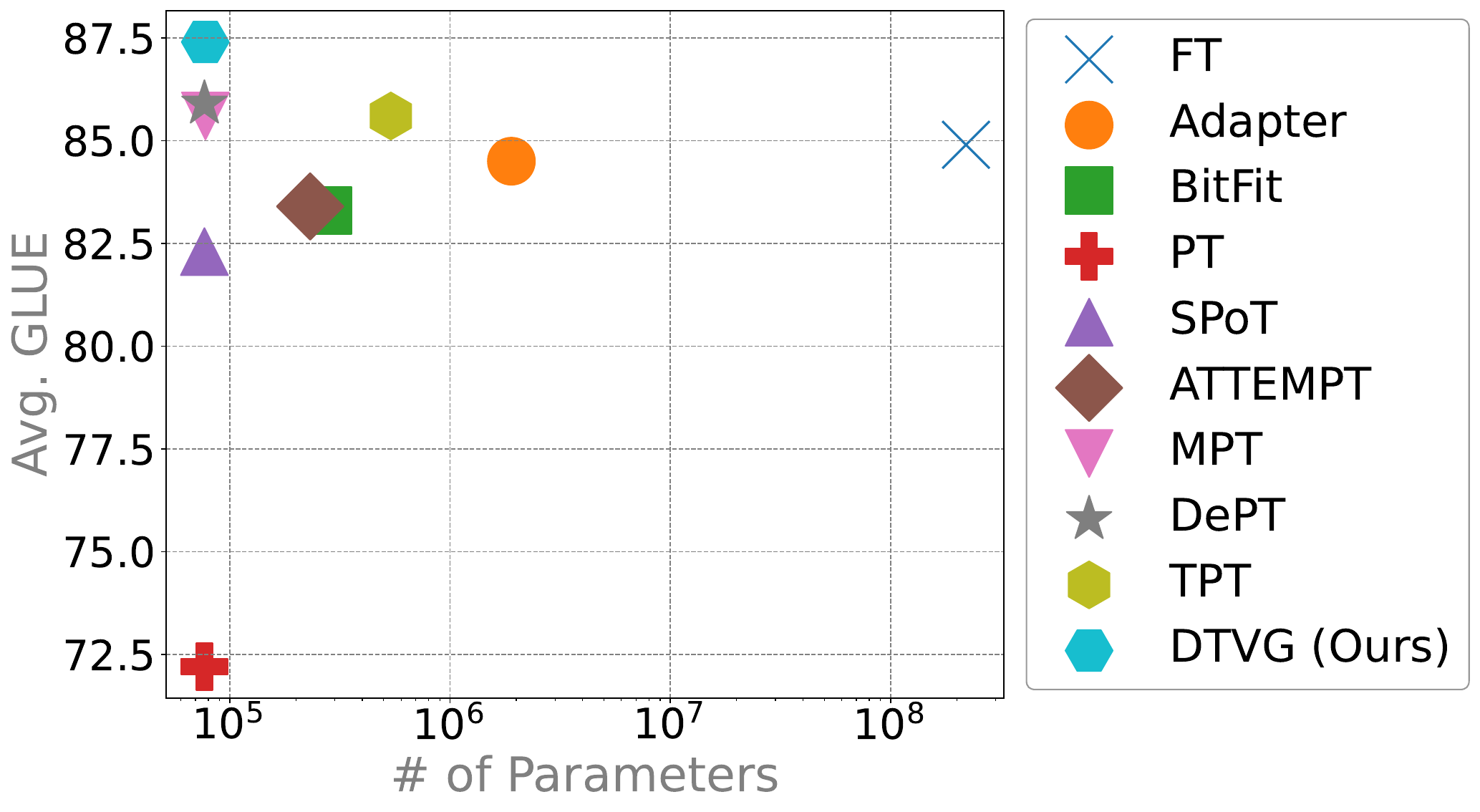} 
    \caption{Performance \& parameter comparison.}
    \label{fig:G_SG}
\end{figure}

We visualize the average score (y-axis) and parameter (x-axis) on the GLUE benchmark across various baselines in Figure~\ref{fig:G_SG}. We observe DTVG surpassing other
baselines and achieving SOTA performance with minimal parameters.

\label{sec:vis_glue_superglue}

\section{Dataset Details}
\label{sec:Dataset}
We use 26 datasets in total from 5 benchmarks. We use GLUE and SuperGLUE benchmarks to test the model’s natural language understanding. MNLI~\cite{williams2018broad}, QNLI~\cite{demszky2018transforming}, QQP~\cite{wang2018glue}, SST2~\cite{socher2013recursive}, RTE~\cite{giampiccolo2007third}, CoLA~\cite{warstadt2019neural}, STS-B~\cite{cer2017semeval}, MRPC~\cite{dolan2005automatically} are derived from GLUE. MultiRC~\cite{khashabi2018looking}, BoolQ~\cite{clark2019boolq}, WiC~\cite{pilehvar2019wic}, WSC~\cite{levesque2012winograd}, and CB~\cite{de2019commitmentbank}, ReCoRD~\cite{zhang2018record} are from SuperGLUE. We use four question-answering datasets from the MRQA 2019 benchmarks, including Natural Questions (NQ)~\cite{kwiatkowski2019natural}, HotpotQA (HQ)~\cite{yang2018hotpotqa}, NewsQA (News)~\cite{trischler2017newsqa}, and SearchQA (SQA)~\cite{dunn2017searchqa}, SQuAD~\cite{rajpurkar2016squad}. WinoGrande (WG)~\cite{sakaguchi2021winogrande}, YelpPolarity (Yelp)~\cite{zhang2015character}, SciTail~\cite{khot2018scitail} and PAWS-Wiki (PAWS)~\cite{zhang2019paws} are from the `other` benchmark to test model’ generalizability across different domains. We also use CommonGen~\cite{DBLP:conf/emnlp/LinZSZBCR20}, E2E~\cite{dusek-etal-2019-semantic}, and WebNLG~\cite{DBLP:conf/inlg/GardentSNP17} sourced from the GEM~\cite{gehrmann-etal-2021-gem} benchmark to test the model's performance on natural language generation. We download all datasets from the huggingface dataset\footnote{\url{https://github.com/huggingface/datasets}}. Table~\ref{tab:Detail_Dateset} lists more details about each dataset. 
\begin{table*}[ht]
\centering
   \begin{subtable}
  \centering
  \tabcolsep=0.1cm
  \small
  \begin{tabular}{l|cc|llll}
\toprule
\bf Dataset & \bf Source & \bf Target     &\bf Benchmark  &\bf Task Type & \bf Domain & \bf Metric \\
\midrule
MNLI  & \checkmark & \checkmark  & GLUE       & Natural Language Inference & Various  & \underline{Accuracy} \\ 
QQP   & \checkmark & \checkmark  & GLUE       & Paraphrase Detection & Social QA & \underline{Accuracy} \& F1 \\
QNLI  & \checkmark & \checkmark  & GLUE (QA)     & Natural Language Inference & Wikipedia  & \underline{Accuracy} \\ 
SST2  & \checkmark  & \checkmark & GLUE       & Sentiment Analysis & Movie Reviews    & \underline{Accuracy}  \\ 
STS-B & $\times$ & \checkmark  & GLUE           & Sentence Similarity & Various & \underline{Pearson} \& Spearman corr. \\ 
MRPC  & $\times$ & \checkmark  & GLUE           & Paraphrase Detection & News & \underline{Accuracy} \& F1 \\ 
RTE   & $\times$ & \checkmark  & GLUE           & Natural Language Inference & News \& Wikipedia & \underline{Accuracy} \\  
CoLA  & $\times$ & \checkmark  & GLUE           & Acceptability & Various & \underline{Matthews corr.} \\ 
ReCoRD      & \checkmark & $\times$ & SuperGLUE & Question Answering (QA) & News & \underline{F1} \& EM \\
MultiRC     & $\times$ & \checkmark & SuperGLUE & Question Answering (QA) & Various & \underline{F1} \& EM \\     
BoolQ       & $\times$ & \checkmark & SuperGLUE & Question Answering (QA) & Wikipedia & \underline{Accuracy} \\ 
WiC         & $\times$ & \checkmark & SuperGLUE & Word Sense Disambiguation & Lexical databases & \underline{Accuracy} \\ 
WSC         & $\times$ & \checkmark & SuperGLUE & Common Sense Reasoning    & Fiction books & \underline{Accuracy} \\ 
CB          & $\times$ & \checkmark & SuperGLUE & Natural Language Inference    & Various & \underline{Accuracy} \\ 
SQuAD       & \checkmark & $\times$ & MRQA 2019 & Question Answering (QA) & Wikipedia & \underline{F1} \& EM \\
NQ & $\times$ & \checkmark & MRQA 2019 & Question Answering (QA) & Wikipedia & \underline{F1} \& EM \\ 
HotpotQA         & $\times$ & \checkmark & MRQA 2019 & Question Answering (QA) & Wikipedia  & \underline{F1} \& EM \\ 
SearchQA         & $\times$ & \checkmark & MRQA 2019 & Question Answering (QA) & Search snippets & \underline{F1} \& EM \\ 
NewsQA           & $\times$ & \checkmark & MRQA 2019 & Question Answering (QA) & News  &  \underline{F1} \& EM \\ 
WinoGrande   & $\times$ & \checkmark & `Other` & Common Sense Reasoning & WikiHow  & \underline{Accuracy} \\ 
YelpPolarity & $\times$ & \checkmark & `Other` & Sentiment Analysis & Yelp reviews & \underline{Accuracy} \\ 
SciTail      & $\times$ & \checkmark & `Other` & Natural Language Inference   & Science exams    & \underline{Accuracy} \\ 
PAWS        & $\times$ & \checkmark & `Other` & Paraphrase Detection & Wikipedia  & \underline{Accuracy} \\ 
WebNLG      & $\times$ & \checkmark & GEM & Data to Text (NLG) & Various    & \underline{Automated Evaluation} \\
E2E         & $\times$ & \checkmark & GEM & Data to Text (NLG) & Restaurant  & \underline{Automated Evaluation} \\
CommonGen   & $\times$ & \checkmark & GEM & Data to Text (NLG) & Commonsense & \underline{Automated Evaluation} \\
\bottomrule
    \end{tabular}  
    \caption{Details about 26 datasets from 5 Benchmarks in total. GLUE (QA) denotes the QNLI derived from the Question Answering Dataset (SQuAD). Lexical databases contain WordNet, VerbNet, and Wiktionary, Search snippets denote question answering from the search engine. Automated Evaluation includes BLEU, Rouge-1, Rouge-2, and Rouge-L. Following \citet{DBLP:conf/iclr/ShiL24}, we use the metric marked with an underline as the primary evaluation metric.}
    \label{tab:Detail_Dateset}    
\end{subtable}
\end{table*}

\section{Implementation Details}
\label{sec:Implementation}
We use PyTorch\footnote{\url{https://pytorch.org/}}, huggingface transformers\footnote{\url{https://github.com/huggingface/transformers}} to implement our method. We validate the effectiveness of DTVG based on the open-source repository~\footnote{\url{https://github.com/AkariAsai/ATTEMPT}}. All of the experiments are conducted with a single GPU with 32 GB of memory. Following~\citet{asai2022attempt}, we use the original T5 checkpoint. We set the batch size for T5-base as 32 for most datasets. We set the batch size to 16 and the gradient accumulation step to 2 for the MRQA benchmark with a long context. Due to the different input lengths of various datasets, we set the maximum token length of 256 for most datasets that have a context of fewer than 200 tokens. We set the maximum token length of 348 for MultiRC and 512 for MRQA datasets. We limit the maximum training data number of YelpPolarity to 100k. We maintain the same hyperparameter settings~\cite{lester2021power} to reinitialize and retrain all tasks, aiming to reconstruct the corresponding soft prompts and task prompt vectors. Similar to~\cite{mahabadi2021parameter}, for datasets lacking publicly available test sets, we use the validation set as the test set or partition it to create separate test and validation sets.

\subsection{Comparison of Task Prompt Vectors and Soft prompt}
\label{sec:TPV_SPoT}
We used the reconstructed soft prompts with the same initialization to compare SPoT~\cite{vu2022spot} and TPV. Specifically, we initialize the target task prompt with the soft prompt that obtained the highest metric score from six source tasks (namely, MNLI, QNLI, QQP, SST-2, ReCoRD, and SQuAD). Note that the difference between the implementations of the two methods SPoT and TPV is only in the task similarity metric. SPoT uses the traditional cosine similarity of soft prompts, while TPV uses Eqn.\ref{eqn:task_sim} to compute task similarity.

\subsection{Full-dataset Transfer}
We set warmup steps to be 500, weight decay to be $1*{10}^{-5}$, and use Adam~\cite{DBLP:journals/corr/KingmaB14} for optimization with a linear learning rate scheduler.

\subsection{Few-shot Adaptation}
In few-shot adaptation experiments, followed by~\cite{mahabadi2021parameter}, we run experiments three times with different random seeds and take the mean of the performance. In each trial, we train 1k steps on the target task for both task prompt vector learning and multi-task prompt transfer stage, which we found to be able to achieve full convergence. We evaluate every 50 steps on the original validation set. For the rest, we report on the original test sets based on the best checkpoint on the validation set.

\subsection{Model Scale}
\label{sec:model_scale}
For model scaling experiments, we set the batch sizes are 100 and 16 for T5-small and T5-large, respectively.

\subsection{Other LLMs}
\label{sec:other_LLMs}
We use Llama-3.2-1B and Llama-3.2-3B to test DTVG's generalizability on other types of LLMs. In our experiment, we use the same 6 source tasks as our main experiments setting on the T5-base and select RTE, CoLA, CB, and WSC as target tasks. We set the length of the soft prompt to 100 for both models and set the batch size to 16 and 4 for Llama-3.2-1B and Llama-3.2-3B, respectively. Compared to the encoder-decoder-based model, we observe that decoder-based autoregressive models need a smaller learning rate, so we set the learning rate of the soft prompt and corresponding scaling term to 0.001.

\subsection{Natural Language Generation}
\label{sec:nlg_tasks}
We select E2E, CommonGen, and WebNLG sourced from the GEM benchmark to evaluate DTVG's performance on natural language generation (NLG) tasks. We use T5-base as the backbone and reuse the task prompt vectors sourced from 6 natural language understanding (NLU) source tasks. We set the maximum 128 token length for both the input and output. We use the target as a simple reference to compute metrics for both PT and DTVG and report the best result on the validation set in Table~\ref{tab:NLG_tasks}.

\subsection{Two Speed Learning Rate}
For the full-dataset transfer setting, we search the learning rate within the set \{3e-1, 4e-1, 5e-1\} for the target task prompt and corresponding scaling term. For the scaling term of the source prompt task vectors, we search the learning rate within the set \{4e-1, 6e-1, 8e-1, 1\}. For few-shot adaptation and others, we set the learning rate of 0.3 for both the target task prompt and corresponding scaling term, and 0.4 for the scaling term of the source task prompt vectors.

\subsection{Prompt Initialization}
We initialized the soft prompt by randomly sampling the top 5000 vocabulary words for all tasks. In both full-dataset transfer and few-shot adaptation experiments, we utilize soft prompt tasks vectors from source tasks by full-dataset prompt tuning. In few-shot adaptation setting, we exclude the corresponding task prompt vectors when adapt to source tasks in GLUE.

\section{MRQA and `Other` Benchmark}
\label{sec:MRQA_other_Benchmark}
As shown in Table~\ref{tab:MRQA_Other}, DTVG realizes significant improvements over the vanilla prompt tuning 
with a 3.7\%, 14.2\% increase on MRQA and `Other` in terms of relative average performance. Compared to other baselines, DTVG also achieves comparable or better performance on MRQA and `other` benchmarks.

\begin{table*}
  \centering
  \tabcolsep=0.08cm
  \small
    \begin{tabular}{ll|ccccc|ccccc}
        \toprule
        & & \multicolumn{5}{|c|}{\bf{MRQA}} & \multicolumn{5}{c}{\bf{Other}} \\
        \midrule
        \multirow{2}{*}{\bf{Method}} & \multirow{2}{*}{ \shortstack{\bf{param} \\ $\backslash$ \bf{task}} } & NQ & HP & SQA & News & \multirow{2}{*}{Avg.} & WG & Yelp & SciTail & PAWS & \multirow{2}{*}{Avg.} \\
        & & (100k) & (72K) & (117K) & (74K) & & (40K) & (100k) & (27K) & (49K) & \\
        \hline
        Finetuning$_{1}$ & \multicolumn{1}{|c|}{220M} & 75.1 & 77.5 & 81.1 & 65.2 & 74.7 & 61.9 & 96.7 & 95.8 & 94.1 & 87.1 \\
        Adapters$_{1}$ & \multicolumn{1}{|c|}{1.9M} & 74.2 & 77.6 & 81.4 & 65.6 & 74.7 & 59.2 & 96.9 & 94.5 & 94.3 & 86.2 \\
        BitFit$_{1}$ & \multicolumn{1}{|c|}{280K} & 70.7 & 75.5 & 77.7 & 64.1 & 72.0 & 57.2 & 94.7 & 94.7 & 92.0 & 84.7 \\
        PT$_{1}$ & \multicolumn{1}{|c|}{76.8K} & 67.9 & 72.9 & 75.7 & 61.1 & 69.4 & 49.6 & 95.1 & 87.9 & 55.8 & 72.1 \\
        LoRA$_{3}$ & \multicolumn{1}{|c|}{3.8M} & 72.4 & 62.3 & 72.5 & 56.9 & 66.0 & 58.2 & 97.1 & 94.7 & 94.0 & 86.0 \\
        SPoT$_{1}$ & \multicolumn{1}{|c|}{76.8K} & 68.2 & 74.8 & 75.3 & 58.2 & 69.1 & 50.4 & 95.4 & 91.2 & 91.1 & 82.0 \\
        ATTEMPT$_{1}$ & \multicolumn{1}{|c|}{232K} & 70.4 & 75.2 & 77.3 & 62.8 & 71.4 & 57.6 & 96.7 & 93.1 & 92.1 & 84.9 \\
        MPT$_{2}$ & \multicolumn{1}{|c|}{77.6k} & 72.0\textsubscript{0.1} & 75.8\textsubscript{0.1} & 77.2\textsubscript{0.1} & 63.7\textsubscript{0.1} & 72.2 & 56.5\textsubscript{0.9} & 96.4\textsubscript{0.0} & 95.5\textsubscript{0.1} & 93.5\textsubscript{0.1} & 85.5 \\
        DePT$_{3}$ & \multicolumn{1}{|c|}{76.8k} & 73.2\textsubscript{0.1} & 76.8\textsubscript{0.3} & 77.6\textsubscript{0.2} & 64.4\textsubscript{0.1} & 73.0 & 59.0\textsubscript{0.2} & 96.8\textsubscript{0.1} & 95.6\textsubscript{0.2} & 93.7\textsubscript{0.1} & 86.3 \\

        \hline
        DTVG (ours) & \multicolumn{1}{|c|}{77.5k} & 73.1\textsubscript{0.1} & 76.7\textsubscript{0.0} & 77.8\textsubscript{0.3} & 64.6\textsubscript{0.1} & 73.1 & 58.0\textsubscript{0.0} & 96.6\textsubscript{0.1} & 97.0\textsubscript{0.1} & 93.7\textsubscript{0.0} & 86.3 \\
        \hline
    \end{tabular}
    
    \caption{Performance on MRQA2019 and `Other` benchmarks. ``param $\backslash$ task`` denotes the number of learnable parameters for each task. $_{1}$ sourced from \cite{asai2022attempt}, $_{2}$ sourced from \cite{DBLP:conf/iclr/WangPKF0K23} and $_{3}$ sourced from \cite{DBLP:conf/iclr/ShiL24}.}
    \label{tab:MRQA_Other}
\end{table*}

\section{Few-shot adaptation On GLUE and SuperGLUE benchmark}

\label{sec:Few-shot GLUE_SuperGLUE}
We compare our method with no transfer baseline PT, one for one baseline DePT, and all for one baseline MPT, and Table~\ref{tab:fs_GLUE_SuperGLUE} shows the evaluation results on GLUE and SuperGLUE benchmarks. Our method can substantially improve the few-shot adaptation results in the most of settings. Specifically, compared to PT, our method on average improves the results across only (k= 4,16,32) shots. Meanwhile, our method also surpasses MPT and DePT, in terms of performance.
\begin{table*}[htp]\small
    \centering
    \resizebox{\textwidth}{!}{
    \begin{tabular}{l|c|ccccccccc|cccccc}
        \toprule
        \multicolumn{2}{c}{} & \multicolumn{9}{c}{\bf{GLUE}} & \multicolumn{6}{c}{\bf{SuperGLUE}} \\
        \hline
        \multicolumn{1}{c}{\bf{Method}} & \multicolumn{1}{|c|}{\bf{k-shot} } & MNLI & QQP & QNLI & SST2 & STS-B & MRPC & RTE & CoLA & \multirow{1}{*}{Avg.} & Multi & Bool & WiC & WSC & CB & \multirow{1}{*}{Avg.} \\
        \hline
        PT & \multirow{4}{*}{\centering 4} & 40.1 & 63.2 & 40.4 & 53.0 & 88.8 & 68.1 & 56.3 & 27.4 & 54.7 & 61.8 & 61.6 & 51.2 & 60.4 & 53.5 & 57.7 \\
        MPT & & 59.4 & 82.0 & 86.2 & 56.5 & 89.1 & 68.1 & 62.6 & 34.8 & 67.3 & 62.2 & 62.2 & 52.9 & 67.3 & 73.6 & 63.6 \\
        DePT & & 44.0\textsubscript{1.1} &77.4\textsubscript{6.7} & 85.8\textsubscript{4.4} & 59.3\textsubscript{3.1} & 84.1\textsubscript{2.7} & 73.5\textsubscript{2.8} & 63.5\textsubscript{2.8}  & 29.3\textsubscript{2.3} & 64.6 & 62.3\textsubscript{1.3} & 62.7\textsubscript{5.4} & 57.5\textsubscript{1.1} & 67.9\textsubscript{0.9}  & 75.0\textsubscript{5.1} & 65.1 \\
        
        Our & & 49.3\textsubscript{1.7} & 87.5\textsubscript{0.7} & 80.2\textsubscript{0.3} & 81.8\textsubscript{1.9} & 87.9\textsubscript{0.5} & 68.1\textsubscript{0.0} & 72.7\textsubscript{0.7} & 22.2\textsubscript{4.5} & 68.7 & 61.4\textsubscript{0.2} & 60.6\textsubscript{1.5} & 59.4\textsubscript{1.7} & 45.2\textsubscript{1.0} & 86.9\textsubscript{1.7} & 62.7 \\
        \hline
        PT & \multirow{4}{*}{\centering 16} & 41.5 & 62.3 & 87.4 & 50.9 & 87.8 & 68.1 & 54.7 & 28.5 & 56.7 & 60.3 & 61.9 & 48.9 & 44.2 & 63.5 & 55.8 \\
        MPT & & 61.6 & 84.7 & 90.6 & 63.2 & 89.1 & 70.1 & 64.8 & 32.1 & 69.5 & 64.5 & 63.3 & 49.8 & 67.3 & 78.6 & 64.7 \\
        
        DePT & & 61.8\textsubscript{2.5} & 80.3\textsubscript{1.3} & 91.2\textsubscript{0.5} & 77.6\textsubscript{6.3} & 87.1\textsubscript{1.7} & 78.1\textsubscript{2.3} & 71.9\textsubscript{1.0} & 27.1\textsubscript{1.7} & 71.9 & 60.6\textsubscript{2.8} & 66.9\textsubscript{4.4} & 59.6\textsubscript{0.7} & 57.7\textsubscript{2.7} & 78.6\textsubscript{4.3} & 
    64.7 \\
        Our & & 58.8\textsubscript{0.6}  & 81.9\textsubscript{1.2} & 89.8\textsubscript{1.1} &  84.6\textsubscript{1.1} & 88.4\textsubscript{0.4} & 86.9\textsubscript{0.4} & 76.8\textsubscript{1.0} & 31.3\textsubscript{2.3} & 74.8 & 61.4\textsubscript{3.1} & 72.3\textsubscript{1.4} & 60.7\textsubscript{0.4} & 67.3\textsubscript{0.0} & 82.1\textsubscript{2.9} & 68.8\\
        \hline
        PT & \multirow{4}{*}{\centering 32} & 37.0 & 62.3 & 56.7 & 50.9 & 87.5 & 68.1 & 54.7 & 23.2 & 55.1 & 59.2 & 61.7 & 52.6 & 67.3 & 67.8 & 61.7 \\
        MPT & & 63.6 & 88.5 & 91.0 & 75.9 & 89.7 & 74.5 & 59.7 & 30.8 & 71.7 & 63.3 & 68.9 & 53.9 & 67.3 & 82.1 & 67.1 \\
        DePT & &     63.3\textsubscript{3.5} & 80.1\textsubscript{0.7} & 91.3\textsubscript{0.5} & 80.4\textsubscript{8.7} & 89.2\textsubscript{0.1} & 81.4\textsubscript{3.3} & 72.7\textsubscript{2.9} & 28.6\textsubscript{2.1} & 73.4 & 60.1\textsubscript{2.7} & 67.2\textsubscript{3.4} & 58.0\textsubscript{0.7} & 63.1\textsubscript{3.6} & 82.1\textsubscript{2.3} & 66.4 \\
        
        Our & & 61.2\textsubscript{0.1} & 85.3\textsubscript{0.8} & 91.2\textsubscript{0.1} & 88.3\textsubscript{1.4} & 83.2\textsubscript{4.7} & 83.1\textsubscript{4.7} & 74.1\textsubscript{2.7} & 29.3\textsubscript{1.5} & 74.5 & 66.3\textsubscript{6.1} & 73.5\textsubscript{1.1} & 60.2\textsubscript{1.0} & 67.3\textsubscript{0.0} & 84.5\textsubscript{1.7} & 70.4 \\
        \hline
    \end{tabular}
    }
    \caption{Few-shot adaptation on GLUE and SuperGLUE benchmark}
    \label{tab:fs_GLUE_SuperGLUE}
\end{table*}

\section{Algorithm Details about DTVG}
\label{sec:DTVG}
We give all the implementation details about DTVG for multi-task prompt tuning on Algorithm~\ref{alg:DTVG}.
\begin{algorithm*}[htbp]  % 使用 algorithm* 环境
    \caption{DTVG}
    \label{alg:DTVG}
    \KwIn{source tasks set $\mathcal{S}=\{s^1, s^2,\dots,s^n\}$, target task $t$, initialization soft prompt parameters $P_{init}$, maximum training steps $N_{\rm max}$ }
    \KwOut{Trained multi-task soft prompt parameters $P_{\rm mix}^{*}$ }
    
    \textbf{Stage 1: Task prompt vector Learning} \;
    Initialize $P_{init}$ for both sources and target task\;
    Boost the posterior probability and obtain their task prompt vectors\;
    \textbf{Stage 2: Multi-task Prompt Transfer} \;
    \For{each iterative $k \gets 1$ \textbf{to} $N_{\rm max}$}{  
    Source Task Grouping: Group a subset of relevant source tasks $\mathcal{S}'$ from $\mathcal{S}$ \;
    Multi-Task Merging: Merge task prompt vectors from $\mathcal{S}' \cup \{t\} $ to get $P_{\rm mix}$ \;
    Boost the posterior probability on target task $t$ with $P_{\rm mix}$
    }
    \KwRet{$P_{\rm mix}^{*}$}
\end{algorithm*}

\section{Details about Source Task Grouping}
\label{sec:STG}
Implementation details about Source Task Grouping for addressing optimization objective~\ref{obj:NP} are presented in Algorithm~\ref{alg:STG}.
\begin{algorithm*}[htbp]
    \caption{Source Task Grouping}
    \label{alg:STG}
    \KwIn{source tasks set $\mathcal{S}=\{s^1, s^2,\dots,s^n\}$, target task $t$}
    \KwOut{selected task group $\mathcal{S}'$}

    \textbf{Step 1: Rank Similarity to Target} \;
    Compute Similarity Ranking list $\Pi$\;
    \textbf{Step 2: Maximize Knowledge Consistency} \;
    Initialize an empty source task group $\mathcal{S}' \gets \emptyset$\;
    \For{each index $\pi^{i}$ from similarity rank list $\Pi$}{
        Let $s^{\pi^{i}}$ be the task corresponding to index $\pi^{i}$\;
        Calculate the contribution of $s^{\pi^i}$ to $\mathcal{S}'$ : $\Delta(\mathcal{S}',s^{\pi^i}) \gets KC(\mathcal{S}' \cup \{s^{\pi^i}\}) - KC(\mathcal{S}')$\;
        \If{$sim(t,s^{\pi^i}) \geq 0$ \textbf{and} $\Delta(\mathcal{S}',s^{\pi^i}) \geq 0$}{
            Add $s^{\pi^i}$ to $\mathcal{S}'$: $\mathcal{S}' \gets \mathcal{S}' \cup \{s^{\pi^i}\}$\;
        }
    }
    \KwRet{$\mathcal{S}'$}
\end{algorithm*}

\section{Source Task Grouping}
\label{sec:TG}
\begin{figure}[t]
    \centering
    \includegraphics[width=\linewidth]{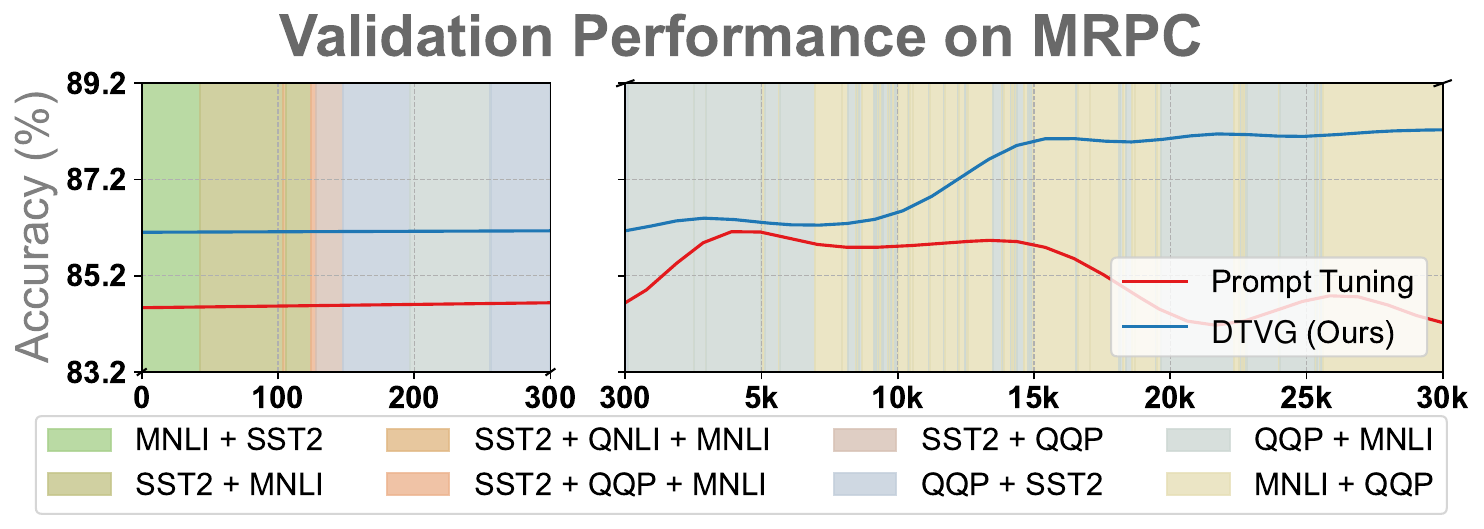}
    \includegraphics[width=\linewidth]{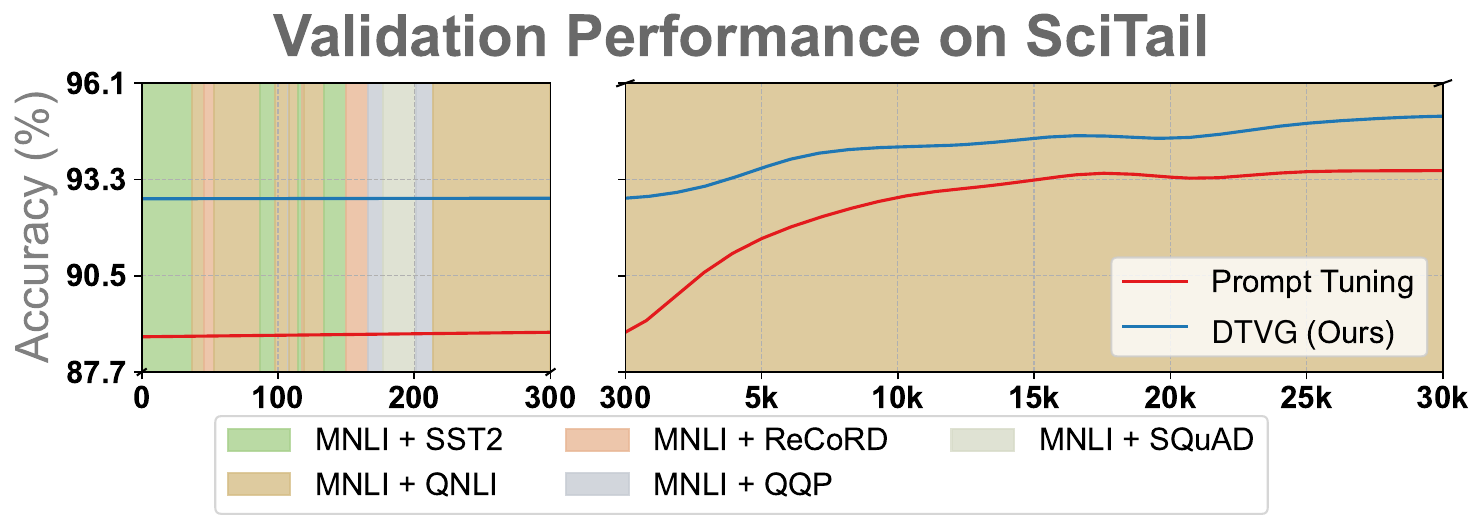}
    \includegraphics[width=\linewidth]{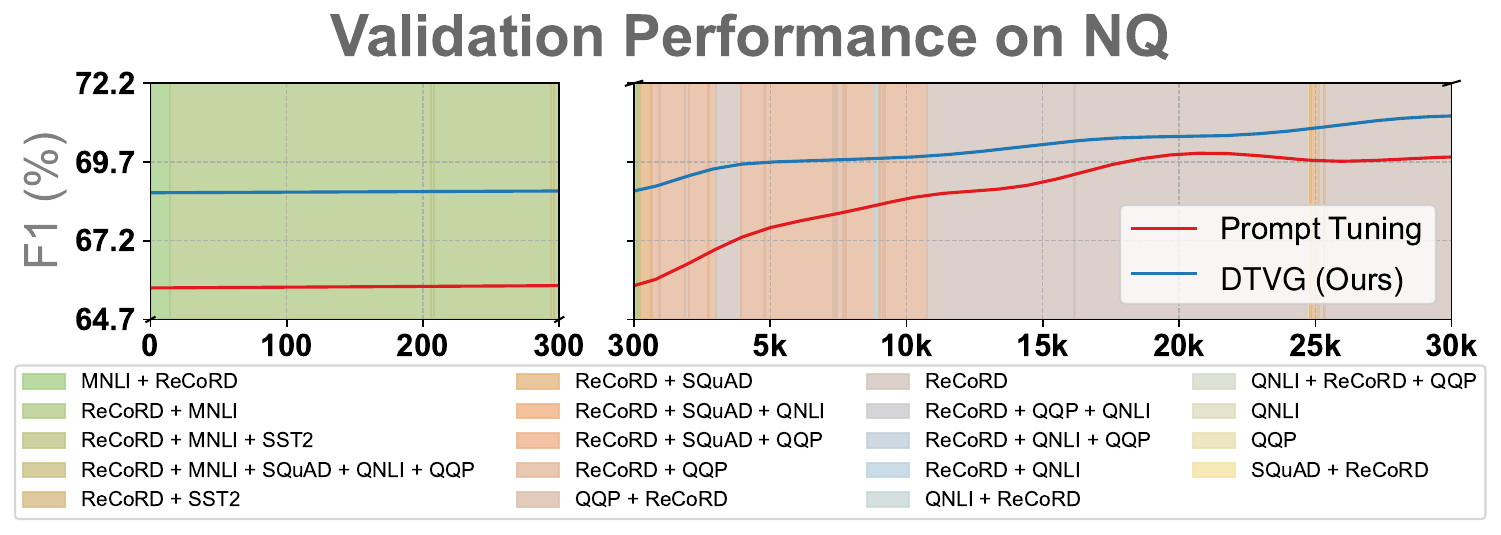}
    \caption{Validation Performance MRPC, SciTail, and NQ with source task grouping.}
    \label{fig:MRPC_SciTail_NQ}
\end{figure}

Figure~\ref{fig:MRPC_SciTail_NQ} demonstrates that DTVG consistently outperforms vanilla prompt tuning in terms of performance across the MRQA, SciTail, and NQ datasets. Meanwhile, DTVG achieves dynamic grouping a appropriate task subset for the different target task. This indicates that DTVG is capable of effectively group a related source task combination tailored to different target tasks, thereby reduce negative transfer.    

\begin{figure}[t]
    \centering 
    \includegraphics[width=\linewidth]{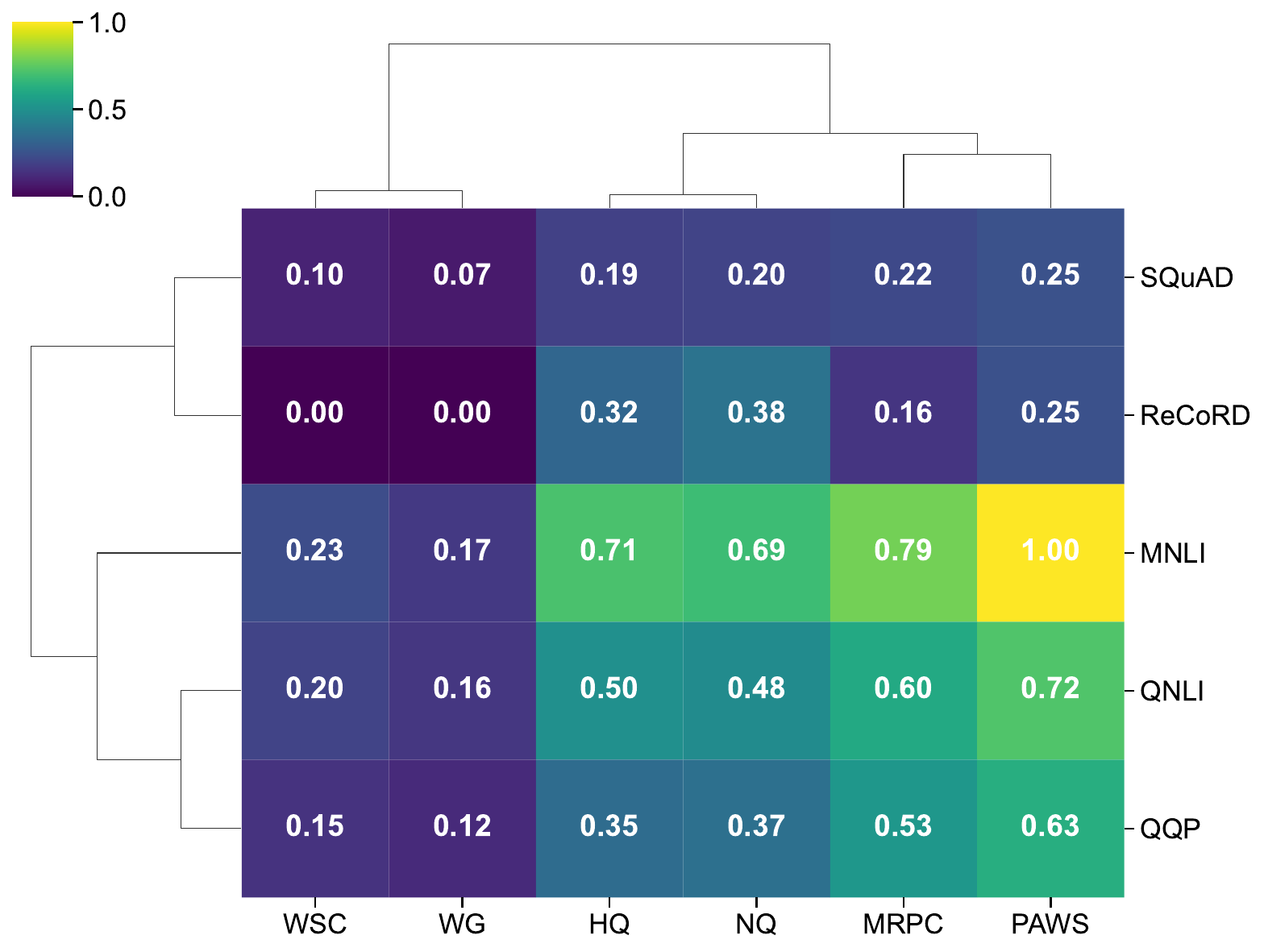}
    \caption{Task similarity of dot product result via TPV. We visualize the task similarity between 5 source tasks and 6 target tasks. We apply min-max normalization to reflect the relative relation among tasks.} 
    \label{fig:Taks Vector}
\end{figure}

\section{Task similarity}
\label{sec:task_sim}
TPV represents the change in parameters after fine-tuning from its initial parameters on a specific task and reflects the specific optimization direction of a task in the weight space. When we fix a unified initialization for all tasks, effectively constraining them to the same weight space, it means that when two TPVs are closer, their optimization directions are more aligned. As a result, when transferring between tasks, there will be fewer conflicts.

\begin{figure*}[htb] 
    \centering 
    \includegraphics[width=\linewidth]{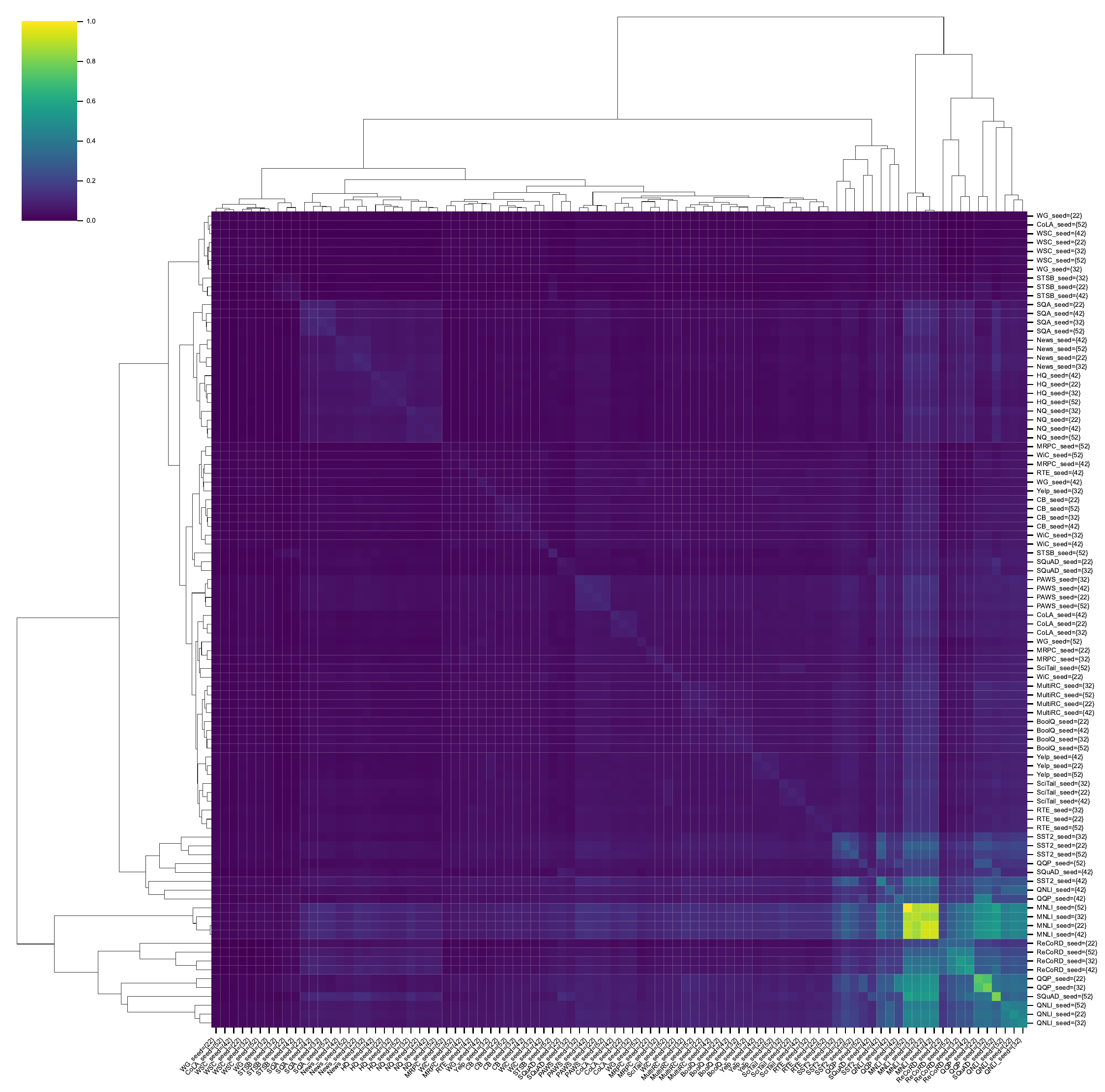}
    \caption{Task similarity visualizations of task prompt vectors. We conduct four experiments with different seeds in $\{22,32,42,52\}$. We apply Min-Max normalization to ensure the relative relationships in the results are maintained.}
    \label{fig:Task_sim}
\end{figure*}

We conduct a case study using 5 source tasks and 6 target tasks with the same initialization to analyze the effectiveness of TPV in capturing the relationships between different tasks. Figure~\ref{fig:Taks Vector} shows the cluster map by computing pairwise task similarity score based on TPV~(Eqn. \ref{eqn:task_sim}). We observe that tasks perceived as similar are clustered together. Specifically, in the source tasks partition, SQuAD and ReCoRD are grouped in the QA cluster. QNLI and QQP belong to QA datasets. This clustering pattern is also observed in the target tasks partition. NQ and HQ are in the QA cluster, MRPC PAWS are Paraphrase Detection, and WSC and WG are Common Sense Reasoning. Furthermore, all target tasks show a consistently high relative task similarity with MNLI, a widely used intermediate task for fine-tuning PLMs~\cite{phang2018sentence}. This highlights the TPV's ability to capture less obvious positive transfer. More details can be found in Figure~\ref{fig:Task_sim}.

\label{sec:TC}

\begin{table*}[ht]
    \centering
    \begin{tabular}{c|cc}
    \toprule
    Method & Test Acc on RTE & Traning samples per second \\
    \hline    
    PT   & 74.1 & 64.2 \\
    DTVG & 86.3 & 58.6 \\
    \hline
    \end{tabular}
    \caption{Test result and training speed on RTE. We use T5-base as backbone}
    \label{tab:Compute_time_cost}
\end{table*}

\section{Computation and Time Costs} 
\label{sec:computation_time_cost}
Dynamically calculating the task combinations during each parameter update does indeed introduce additional time and computation costs during training. However, this computation does not involve gradients, so it ultimately does not lead to a significant increase in time and computation burden. We visualize the result and training speed on RTE in Table~\ref{tab:Compute_time_cost}. We observe that DTVG demonstrates a 16.5\% improvement in performance while incurring only a 8.7\% decrease in training speed compared to prompt tuning.

\end{document}